\documentclass[10pt,journal,compsoc]{IEEEtran}

\usepackage{color}
\usepackage{caption}
\usepackage{tcolorbox}
\usepackage{times}
\usepackage{helvet}
\usepackage{courier}
\usepackage{amsmath}
\usepackage{amssymb}
\usepackage{amsthm}
\usepackage{graphicx}
\usepackage{array}
\usepackage{booktabs}
\usepackage{amsopn}
\usepackage{amsmath,bm}
\usepackage{algorithm}
\usepackage{algorithmic}
\usepackage{enumerate}
\usepackage{color}
\usepackage{multirow}
\usepackage{bbding}
\usepackage{threeparttable}
\usepackage{threeparttable} 
\usepackage{dsfont} 
\usepackage{makecell} 
\usepackage{url}            
\usepackage{CJKutf8}
\usepackage{subcaption}
\usepackage{pifont}

\newcommand{\eg}{\emph{e.g.}}
\newcommand{\etal}{\emph{et al.}}
\newcommand{\ie}{\emph{i.e.}}
\newcommand{\etc}{\emph{etc}}

\newcommand{\qileft}{[\kern-0.15em[}
\newcommand{\qiLeft}{\left[\kern-0.4em\left[}
\newcommand{\qiright}{]\kern-0.15em]}
\newcommand{\qiRight}{\right]\kern-0.4em\right]}

\renewcommand{\O}{\mathcal{O}}

\graphicspath{{figs/}}
\begin{document}
	\begin{CJK}{UTF8}{gbsn}
	\title{A Survey on Transformer Compression}
		\author{Yehui Tang, Yunhe Wang, Jianyuan Guo, Zhijun Tu, Kai Han, Hailin Hu, and Dacheng Tao~\IEEEmembership{Fellow,~IEEE}
			\IEEEcompsocitemizethanks{
				\IEEEcompsocthanksitem Yehui Tang, Jianyuan Guo, Zhijun Tu, Kai Han, Hailin Hu, and Yunhe Wang are with Huawei Noah's Ark Lab. E-mail: yunhe.wang@huawei.com.
				\IEEEcompsocthanksitem Dacheng Tao is with the School of Computer Science, in the Faculty of Engineering, at The University of Sydney, 6 Cleveland St, Darlington, NSW 2008, Australia. E-mail: dacheng.tao@sydney.edu.au.
				\IEEEcompsocthanksitem Corresponding to Yunhe Wang and Dacheng Tao.
			}
		}
	
	\IEEEtitleabstractindextext{%
\begin{abstract}
Transformer plays a vital role in the realms of natural language processing (NLP) and computer vision (CV), specially for constructing large language models (LLM) and large vision models (LVM).  Model compression methods reduce the memory and computational cost of Transformer, which is a necessary step to implement large language/vision models on practical devices. Given the unique architecture of Transformer, featuring alternative attention and feedforward neural network (FFN) modules, specific compression techniques are usually required. The efficiency of these compression methods is also paramount, as retraining large models on the entire training dataset is usually  impractical. This survey provides a comprehensive review of recent compression methods, with a specific focus on their application to  Transformer-based models. The compression methods are primarily categorized into pruning, quantization, knowledge distillation, and efficient architecture design (Mamba, RetNet, RWKV, \etc). In each category, we discuss compression methods for both language and vision tasks, highlighting common underlying principles. Finally, we  delve into the relation between various  compression methods, and discuss  further directions in this domain.
\end{abstract}
		
}
	
	\maketitle
	
	\IEEEdisplaynontitleabstractindextext
	
	\IEEEpeerreviewmaketitle
	
	\IEEEraisesectionheading{\section{Introduction}\label{Sec:Introduction}}
\IEEEPARstart{D}{eep} neural networks have become indispensable in numerous artificial intelligence applications, with architectures encompassing diverse formulations, such as multilayer perceptron (MLP), convolutional neural network (CNN), recurrent neural network (RNN), long short-term memory (LSTM), Transformers, \etc. In recent times, Transformer-based models have emerged as the prevailing choice across various domains, including both natural language processing (NLP) and computer vision (CV) domains. Considering their strong scalability,  most of the large models with over billions of parameters are based on the Transformer architecture, which are  considered  as  foundational elements for artificial general intelligence (AGI)~\cite{touvron2023llama,gpt4,Radford2019LanguageMA,brown2020language,zeng2021pangu,ren2023pangu}.

While large models have demonstrated significant capabilities, their exceptionally vast sizes pose challenges for practical development.   For instance, the GPT-3 model has   175 billion parameters  and demands  about 350GB memory model storage (float16). The sheer volume of parameters and the associated computational expenses necessitate devices with exceedingly high memory and computational capabilities. Directly deploying such models will incur substantial resource costs and contributes significantly to carbon dioxide emissions. Moreover, on edge devices like mobile phones, the development of these models becomes impractical due to the limited storage and computing resources of such devices.

 \begin{figure*}[t]
	\small
	\centering
	\includegraphics[width=1.0\linewidth]{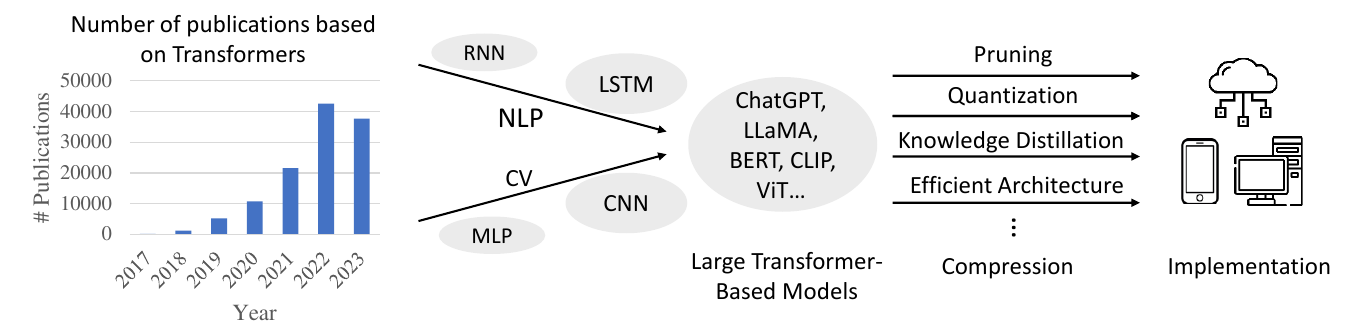}
	\caption{Transformer-based models have emerged as the predominant architectures in both natural language processing (NLP) and computer vision (CV) domains, resulting in a surge in publications. As these models tend to possess substantial dimensions, it becomes imperative to compress their parameters and streamline computational redundancies. This compression is essential for facilitating efficient implementation on practical platforms, ensuring the feasibility of deploying Transformer models in real-world applications.}
	\label{fig:sum}
\end{figure*} 

Model compression is  an effective strategy for mitigating the development costs associated with Transformer models. This approach, grounded in the principle of  reducing redundancy, encompasses various categories, including pruning, quantization, knowledge distillation, efficient architecture design, \etc. Network pruning directly removes redundant components, such as blocks, attention heads, FFN layers, and  individual parameters. Diverse sub-models can be derived by employing different pruning granularity and pruning criteria. Quantization reduces the development cost by representing model weights and intermediate features with lower bits. For example, when quantizing a full-precision model (float32) into 8-bit integers, the memory cost can be reduced by a factor of four. According the computational process, it can be divided into post-training quantization(PTQ) or quantization-aware training (QAT), in which the former  incurs only limited training costs and is more efficient for large models.
Knowledge distillation serves as a training strategy, and transfers knowledge from a large model (teacher) to a smaller model (student). The student mimics the behavior of the teacher by emulating the model's output and intermediate features. Notably, for advanced models like GPT-4, accessible only through APIs, their generated instructions and explanations can also guide the learning of the student model~\cite{jiang2023lion,li2022explanations}.In addition to obtaining models from predefined large models, some methods yield efficient architectures by directly reducing the computational complexity of attention modules or FFN modules. Combining different methods enables extreme compression. For instance, Han et al.~\cite{han2015deep} combined network pruning, quantization, and Huffman coding to achieve an impressive 49$\times$ compression rate on a conventional VGGNet~\cite{simonyan2014very}.

Regarding Transformer models, their compression strategies exhibit distinct characteristics. Unlike other architectures such as CNN or RNN, the Transformer features a unique design with alternative attention and FFN modules. The former captures the global information by calculating the attention map over different tokens while the latter extracts information from each token respectively. This specific architecture can enable a tailored compression strategy for optimal compression rates. What's more, the efficiency of compression method becomes especially important for such large models. Due to the high computational cost of large model, it is usually unaffordable to retrain the whole model on the original training set. Some training-efficient methods like post-training compression are preferable.  

In this survey, we aim to comprehensively investigate how to compress these  Transformer models (Figure~\ref{fig:sum}), and categorize the methods by  quantization, knowledge distillation, pruning, efficient architecture design, \etc. In each category, we investigate the  compression methods for NLP and CV domains, respectively. Table~\ref{tab:overview} summarizes the main compression categories and lists representative method suitable for large Transformer models. Though NLP and CV are usually treated as very different domains,  we observe that their models compression methods actually share the similar principles. Finally, we discuss the relationship between different compression methods and outline some future research directions.


The rest of the paper is organized as follows. Section~\ref{sec:trans} introduces the fundamental concept of Transformers. Following this, Section~\ref{sec:pres} provides an in-depth discussion on compression methods that preserve the architecture, encompassing quantization and knowledge distillation—techniques that maintain the model's  architecture. Section~\ref{sec:apt} delves further into architecture-preserving compression, including pruning and efficient architecture design. Additional Transformer compression methods are explored in Section~\ref{sec:other}. Finally, Section~\ref{sec:con} draws conclusions on the compression methods and discusses future research directions.





\section{Concept of Transformer} 

\label{sec:trans}

The Transformer architecture is firstly proposed to tackle tasks like machine translation~\cite{vaswani2017attention}. A standard Transformer architecture contains main blocks,  multi-head attention (MHA)  and  feed-forward networks (FFN). The attention is formulated as  
\begin{equation}
	\mathrm{Attention}(Q, K, V) = \mathrm{softmax}(\frac{QK^T}{\sqrt{d}})V,
\end{equation}
where $Q$, $K$, $V$ are query, key and value matrix, respectively. $d$ is the feature's dimension. The multi-head attention jointly extracts information from diverse subspaces, which is the concatenation of different heads, 
\begin{equation}
	\label{eq:mulhead}
	\small
	\begin{aligned}
		&\mathrm{MultiHead}(Q, K, V) = \mathrm{Concat}(\mathrm{head_1}, ..., \mathrm{head_h})W^O,\\
		&\text{where}~\mathrm{head_i} = \mathrm{Attention}(QW^Q_i, KW^K_i, VW^V_i).\\
	\end{aligned}
\end{equation}
$W^Q$, $W^K$, $W^V$, $W^O$ are the corresponding parameter matrices. The FFN module transforms  features from each token independently. It is usually constructed by stacking two FC layers with activation functions, 
\begin{equation}
	\label{eq:ffn}
	\mathrm{FFN}(x)=\phi(xW_1 + b_1) W_2 + b_2,
\end{equation}
where $x$ is input feature, and  $\phi$ is activation function (\eg, GELU). $W_1$, $W_2$, $b_1$, $b_2$ are the weight and bias parameters in FC layers. The MHA and FFN module are stacked alternatively to construct the whole model.

The Transformer architecture has strong scalability and so can be used to construct extremely large models with several billion or trillion parameters. It  supports most of the predominant large models in NLP, CV and multiple modality domains. For example, the well-known large language models (\eg, GPT-series~\cite{brown2020language,gpt4}, LLaMA~\cite{touvron2023llama},Pangu~\cite{zeng2021pangu,ren2023pangu}) are its decoder-only variants. By simply splitting an image into multiple patches, it can be used to tackle vision tasks~\cite{Dosovitskiy2020AnII,chen2021pre,bai2023sequential}. The multiple model like CLIP~\cite{radford2021learning}, BLIP~\cite{li2022blip}, LLaVA~\cite{liu2023visual} also use Transformer as the backbones. 


\begin{table*}[htb]
	\centering
	\renewcommand\arraystretch{1.0}
	\caption{Representative compression method for Transformer models.}
	\label{tab:overview}
	\footnotesize
	\setlength{\tabcolsep}{3.5pt}{
		\begin{tabular}{c|c|c|c|c}
			\toprule[1.5pt]
			Category & Sub-category & Method & Highlights & Publication \\
			\hline \hline

			\multirow{6}{*}{{Quantization}} & \multirow{3}{*}{NLP}  &SmoothQuant~\cite{xiao2023smoothquant}  & Training-free, smooth outliers, equivalent transformation & ICML 2023  \\
			& &OmniQuant~\cite{shao2023omniquant} & Weight clipping, learnable transformation, block-wise & Arxiv 2023  \\
			&	& QLoRA~\cite{dettmers2023qlora} & Parameter-efficient fine-tuning, memory management & Arxiv 2023  \\  \cline{2-5}
			& \multirow{3}{*}{CV}  & PTQ-ViT~\cite{liu2021post}  & Self-attention preservation, mixed-precision & NeurIPS 2021 \\
			& &FQ-ViT~\cite{lin2021fq}  & Fully-quantized, log2 quantization, power-of-two factor & IJCAI 2022  \\
			& &OFQ~\cite{liu2023oscillation} & Confidence-guided annealing, query-key reparameterization  & ICML 2023  \\  \hline \hline
			\multirow{6}{*}{{Knowledge Distillation}} & \multirow{3}{*}{NLP}  & DistilBERT~\cite{sanh2019distilbert} & Small version of BERT, trained with logits of the teacher & NeurIPS 2019  \\
			& & MiniLM~\cite{wang2020minilm} & Mimicking attention distribution and value-relation of teacher & NeurIPS 2020  \\
			&	& Lion~\cite{jiang2023lion} & Adversarial distillation: imitation, discrimination, generation & EMNLP 2023 \\  \cline{2-5}
			& \multirow{3}{*}{CV} & DeiT~\cite{Touvron2020TrainingDI} & Hard labels, novel distillation token in ViTs & ICML 2021  \\
			& & TinyViT~\cite{wu2022tinyvit} & Large-scale pretraining data, encoded data augmentation & ECCV 2022  \\
			& & ManifoldKD~\cite{hao2022learning} & Patch-level, batch-level manifold information & NeurIPS 2022  \\  \hline \hline
						\multirow{5}{*}{{Pruning}} & \multirow{3}{*}{NLP}  
			&LLM Pruner~\cite{Ma2023LLMPrunerOT} & Structured, coupled-module identification & NeurIPS 2023 \\
			&	& Sheared LLaMA~\cite{Xia2023ShearedLA} & Structured, pre-defined target, dynamic data loading & NeurIPS 2023  \\ 
			&	&Dynamic Context Pruning~\cite{Anagnostidis2023DynamicCP}  & Sigmoid-based context selection, KV-cache aware & NeurIPS 2023  \\ \cline{2-5}
			& \multirow{3}{*}{CV}  & ViT-Slim~\cite{Chavan2022VisionTS}  & Structured, single-shot architecture search & CVPR 2022 \\
			& &Patch Sliming~\cite{Tang2021PatchSF} & Top-down unimportant patch removing & CVPR 2022  \\
			& &X-pruner~\cite{Yu2023XPrunerEP} & Structured, class-aware, layer-wise fully differentiable pruning & CVPR 2023  \\  \hline \hline
			\multirow{6}{*}{{Efficient Architecture}} & \multirow{3}{*}{NLP} & PaLM~\cite{chowdhery2022palm} & SwiGLU activation in FFN, densely activated & JMLR 2023  \\
			& & RetNet~\cite{sun2023retentive} & Training parallelism, low-cost inference, parallel, recurrent & Arxiv 2023  \\
			& & Reformer~\cite{kitaev2020reformer} & Efficient attention, locality-sensitive hashing & ICLR 2020  \\  \cline{2-5}
			& \multirow{3}{*}{CV}  & Swin~\cite{liu2021swin} & Hierarchical structures, shifted local window attention & ICCV 2021  \\
			& & MetaFormer~\cite{yu2022metaformer} & Non-parametric pooling as basic token mixing & CVPR 2022  \\
			& & MLP-Mixer~\cite{tolstikhin2021mlp} & Architecture based exclusively on multi-layer perceptrons & NeurIPS 2021  \\  
			\bottomrule[1.5pt]
		\end{tabular}
	}
\end{table*}

\section{Architecture Preserved Compression} 
\label{sec:pres}

\subsection{Quantization}
\subsubsection{Overview of Quantization}

\begin{figure*}[t]
	\centering
	\includegraphics[width=0.95\linewidth]{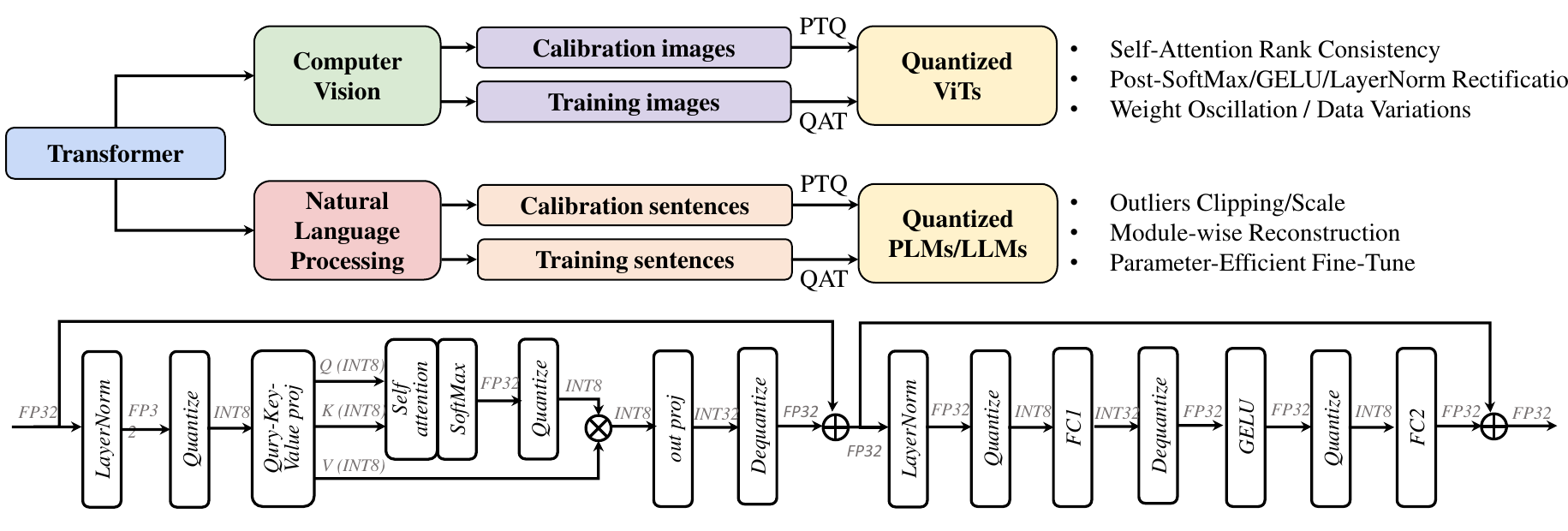}
	\caption{The overview of quantization for Transformers. The top demonstrates the different problems that are addressed in existing works for computer vision and natural language processing, and the bottom shows a normal INT8 inference process of a standard Transformer block.}
	\label{quant_overview}
	\vspace{-2mm}
\end{figure*}

Quantization is a crucial step for deploying  Transformers on various devices, especially GPUs and NPUs, that have specialized circuits for low-precision arithmetic. During the quantization process as shown in Equation~\ref{quant}, a floating-point tensor $x$ is converted to the integer one $x_{int}$ with corresponding quantization parameters (scale factor $s$ and zero point $z$), then the integer tensor $x_{int}$ could be quantized back to floating-point $x_{quant}$ but causes some precision error compared with the original $x$, 
\begin{equation}
	\begin{aligned}
		x_{int}  & = \textrm{Clamp}(\lfloor x/s \rceil +z, 0, 2^b-1), \\
		x_{quant} & = s(x_{int}-z),
		\label{quant}
	\end{aligned}
\end{equation}
where $b$ denotes the bit-width, $\lfloor \cdot \rceil$ represents the rounding function and `Clamp' clips the values that exceed the given range. For the matrix multiplication, the weight $\textrm{w}$ adopts symmetrical quantization with zero point $z_w=0$, and the input embedding tensor $e$ is quantized with unsymmetrical quantization, as shown in Equation~\ref{quant_matmul}:
\begin{equation}
	\begin{aligned}
		y & = \textrm{MatMul}(e, \textrm{w}) \approx \textrm{MatMul}(e_{quant}, \textrm{w}_{quant}) \\
		& = \textrm{MatMul}(s_e(e_{int}-z_e), s_w \textrm{w}_{int}) \\
		& = s_es_w\textrm{MatMul}(e_{int}, \textrm{w}_{int}) + C,
		\label{quant_matmul}
	\end{aligned}
\end{equation}
where $s_w, s_e, z_e$ are quantization parameters of weights and input embedding, and $e_{int}$ and $\textrm{w}_{int}$ are the integer input and weights, which are calculated by Equation~\ref{quant}. $C$ could be pre-computed with $s_e, z_e, s_e$ and $\textrm{w}_{int}$. Thus the floating-point multiplication could be accelerated with efficient integer multiplication in the inference. To minimize the performance degradation of quantized models, different optimization methods have been proposed and they can be divided into two categories: (1) \textbf{Post-training quantization} (PTQ)~\cite{liu2021post, yuan2022ptq4vit,lin2021fq,ding2022towards,lit2022auto, liu2023noisyquant, li2023repq} ,  mainly focuses on optimizing the quantization parameters of weights and activations with a few unlabeled calibration data, and some of the latest methods also explore adaptive rounding for weight quantization. (2) \textbf{Quantization-aware training} (QAT)~\cite{wang2022quantformer,li2022q,li2022q_diff, chen2023data, li2023vit, frumkin2023jumping, liu2023oscillation, xiao2023patch, li2022patch, li2023psaq, xu2023q, huang2023variation},  inserts the quantization nodes into networks and conducts training with complete training data, where all the weights and quantization parameters are optimized together. 
In this section, we systematically introduce the research of model quantization on Transformer-based vision models and large language models, as shown in Figure~\ref{quant_overview}.

\begin{table}[t]
		\centering
		\caption{Comparison of different PTQ and QAT methods for transformer-based vision models. W/A denotes the bit-width of weight and activation and the results show the top-1 accuracy on ImageNet-1k validation set. * represents for mixed-precision. }
		\label{Quant4CV}
		\vspace{-2mm}
		\setlength{\tabcolsep}{4.3pt}
		\scalebox{0.88}{
			\begin{tabular}{l  c | c c c c c}
				\toprule[0.1em]
				\textbf{Method of PTQ} & \textbf{W/A (bit)} & \textbf{ViT-B} & \textbf{DeiT-S} & \textbf{DeiT-B} & \textbf{Swin-S} & \textbf{Swin-B} \\
				\midrule
				\midrule
				Full Precision & 32/32 & 84.54 & 79.85 & 81.80 & 83.23 & 85.27 \\
				\midrule
				PTQ-ViT~\cite{liu2021post} & 8*/8*  & 76.98 & 78.09 & 81.29 & - & - \\
				PTQ4ViT~\cite{yuan2022ptq4vit} & 8/8  & 84.54 & 79.47 & 81.48 & 83.10 & 85.14 \\
				FQ-ViT~\cite{lin2021fq} & 8/8 & 83.31 & 79.17 & 81.20 & 82.71 & 82.97 \\
				APQ-ViT~\cite{ding2022towards} & 8/8  & 84.26 & 79.78 & 81.72 & 83.16 & 85.16\\
				NoiseQuant~\cite{liu2023noisyquant} & 8/8  & 84.22 & 79.51 & 81.45 & 83.13 & 85.20 \\
				\midrule
				PTQ-ViT~\cite{liu2021post} &6*/6*  & 75.26 & 75.10 & 77.47 & - & - \\
				PTQ4ViT~\cite{yuan2022ptq4vit} & 6/6  & 81.65 & 76.28 & 80.25 & 82.38 & 84.01 \\
				APQ-ViT~\cite{ding2022towards} & 6/6  & 82.21 & 77.76 & 80.42 & 84.18 & 85.60\\
				NoiseQuant~\cite{liu2023noisyquant} & 6/6  & 82.32 & 77.43 & 80.70 & 82.86 & 84.68\\
				RepQ-ViT\cite{li2023repq} & 6/6  & 83.62 & 78.90 & 81.27 & 82.79 & 84.57\\
				\midrule
				APQ-ViT~\cite{ding2022towards} & 4/8  & 72.63 & 77.14 & 79.55 & 80.56 & 81.94\\
				\midrule
				PTQ-ViT~\cite{liu2021post} & 4*/4*  & - & - & 75.94 & - & - \\
				APQ-ViT~\cite{ding2022towards} & 4/4 & 41.41 & 43.55 & 67.48 & 83.16 & 85.16 \\
				RepQ-ViT\cite{li2023repq} & 4/4  & 68.48 & 69.03 & 75.61 & 79.45 & 78.32\\
				\midrule
				\textbf{Method of QAT} & \textbf{W/A (bit)} & \textbf{DeiT-T} & \textbf{DeiT-S} & \textbf{DeiT-B} & \textbf{Swin-T} & \textbf{Swin-S} \\
				\midrule
				\midrule
				Full Precision & 32/32 & 72.20 & 79.85 & 81.80 & 81.20 & 83.23  \\
				\midrule
				I-ViT~\cite{li2023vit} & 8/8 & 72.24 & 80.12 & 81.74 & 81.50 & 83.01 \\
				\midrule
				Q-ViT~\cite{li2022q_diff} & 4/4 & 72.79 & 80.11 & - & 80.59 & - \\
				AFQ-ViT~\cite{li2022q_diff} & 4/4 & - & 80.90 & 83.00 & 82.50 & 84.40 \\
				Quantformer~\cite{wang2022quantformer} & 4/4 & 69.90 & 78.20 & 79.70 & 78.30 & 81.00 \\
				OFQ~\cite{liu2023oscillation} & 4/4 & 75.46 & 81.10 & - & 81.88 & - \\
				VVTQ~\cite{huang2023variation} & 4/4 & 74.71 & - & - & 82.42 & - \\
				\midrule
				Q-ViT~\cite{li2022q_diff} & 3/3 & 69.62 & 78.08 & - &  79.45 & - \\
				AFQ-ViT~\cite{li2022q_diff} & 3/3 & - & 79.00 & 81.00 & 80.90 & 82.70 \\
				Quantformer~\cite{wang2022quantformer} & 3/3 & 65.20 & 75.40 & 78.30 & 77.40 & 79.20 \\
				OFQ~\cite{liu2023oscillation} & 3/3 & 72.72 & 79.57 & - & 81.09 & - \\
				\midrule
				AFQ-ViT~\cite{li2022q_diff} & 2/2 & - & 72.10 & 74.20 & 74.70 & 76.90 \\
				Quantformer~\cite{wang2022quantformer} & 2/2 & 60.70 & 65.20 & 73.80 & 74.20 & 76.60 \\
				OFQ~\cite{liu2023oscillation} & 2/2 & 64.33 & 75.72 & - & 78.52 & - \\
				\bottomrule[0.1em]
		\end{tabular}}
		\vspace{-2mm}
	\end{table}

\begin{figure*}[tb]
	\begin{minipage}[t]{0.25\linewidth}
		\centering
		\includegraphics[width=0.9\linewidth]{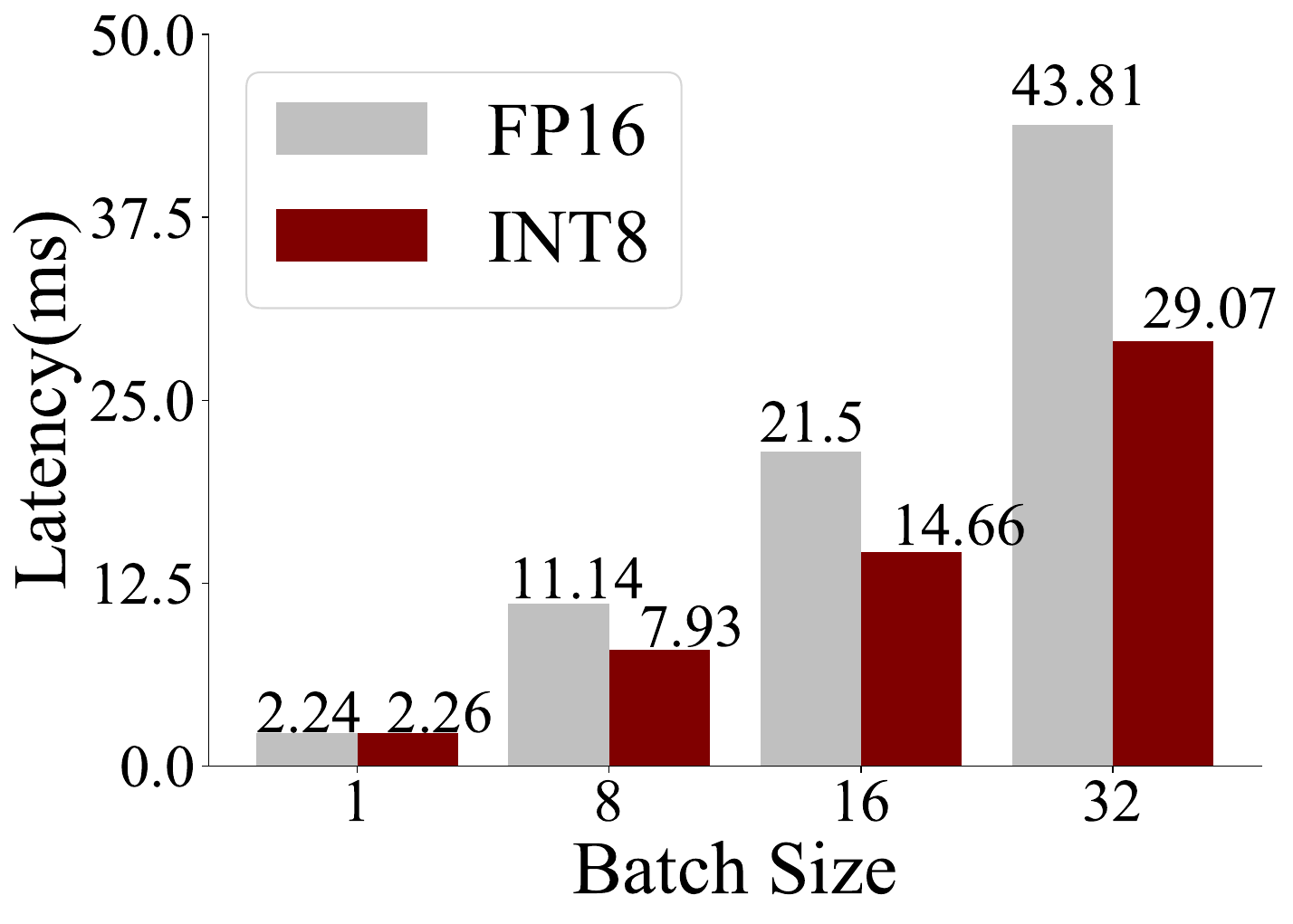}
		\subcaption{ViT-B\_16}
	\end{minipage}%
	\begin{minipage}[t]{0.25\linewidth}
		\centering
		\includegraphics[width=0.9\linewidth]{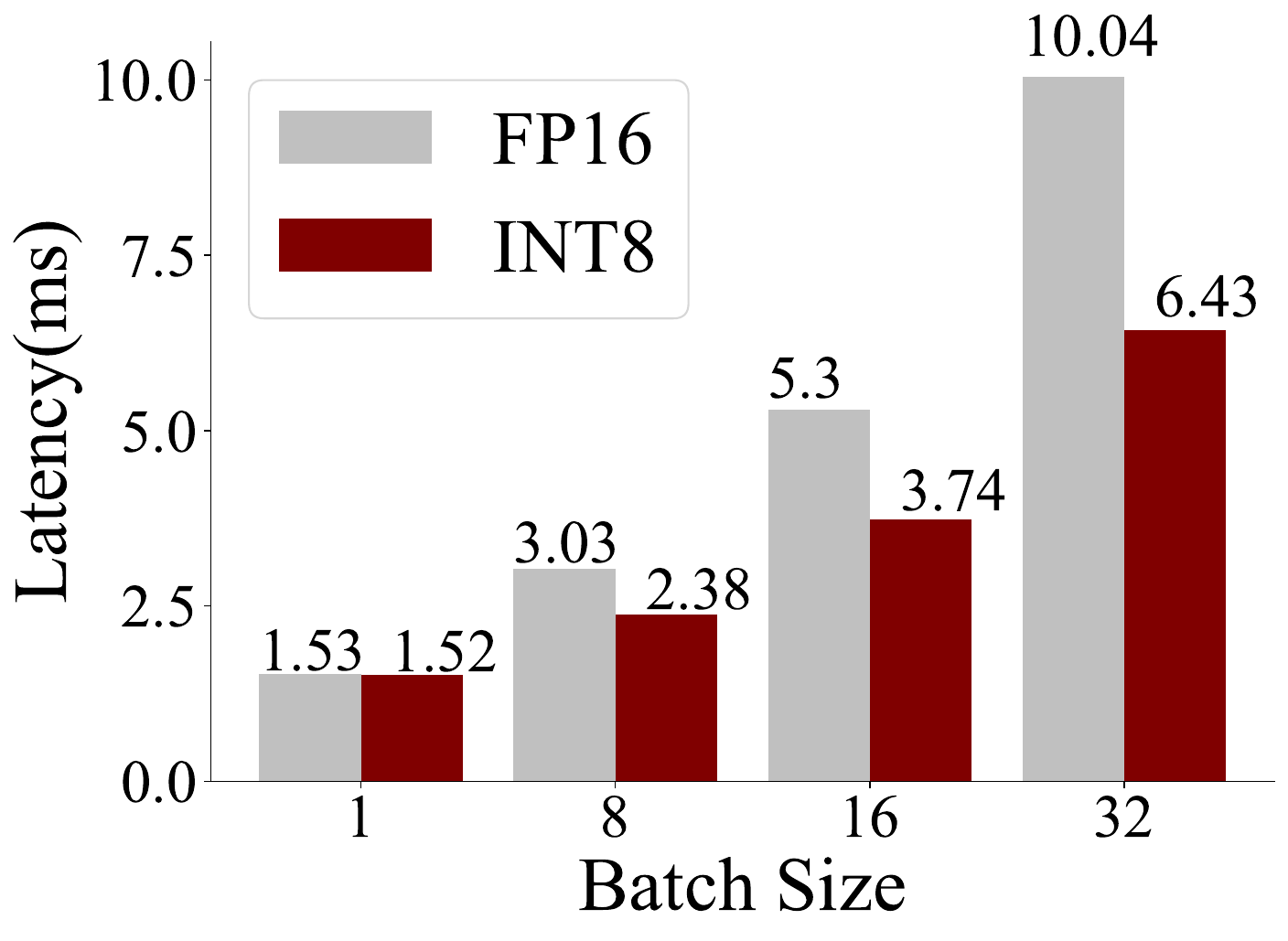}
		\subcaption{ViT-B\_16-224}
	\end{minipage}
	\begin{minipage}[t]{0.25\linewidth}
		\centering
		\includegraphics[width=0.9\linewidth]{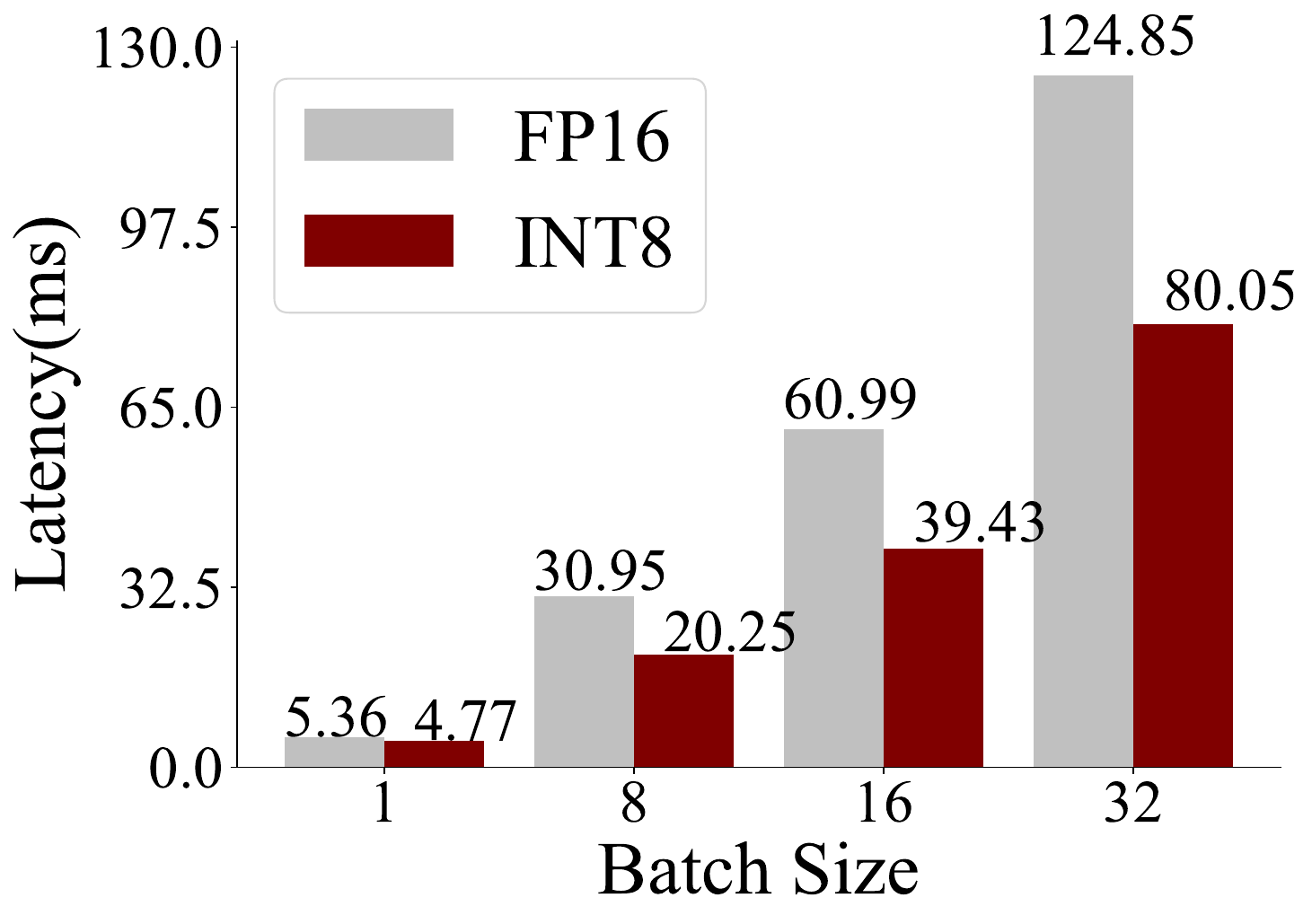}
		\subcaption{ViT-L\_16}
	\end{minipage}%
	\begin{minipage}[t]{0.25\linewidth}
		\centering
		\includegraphics[width=0.9\linewidth]{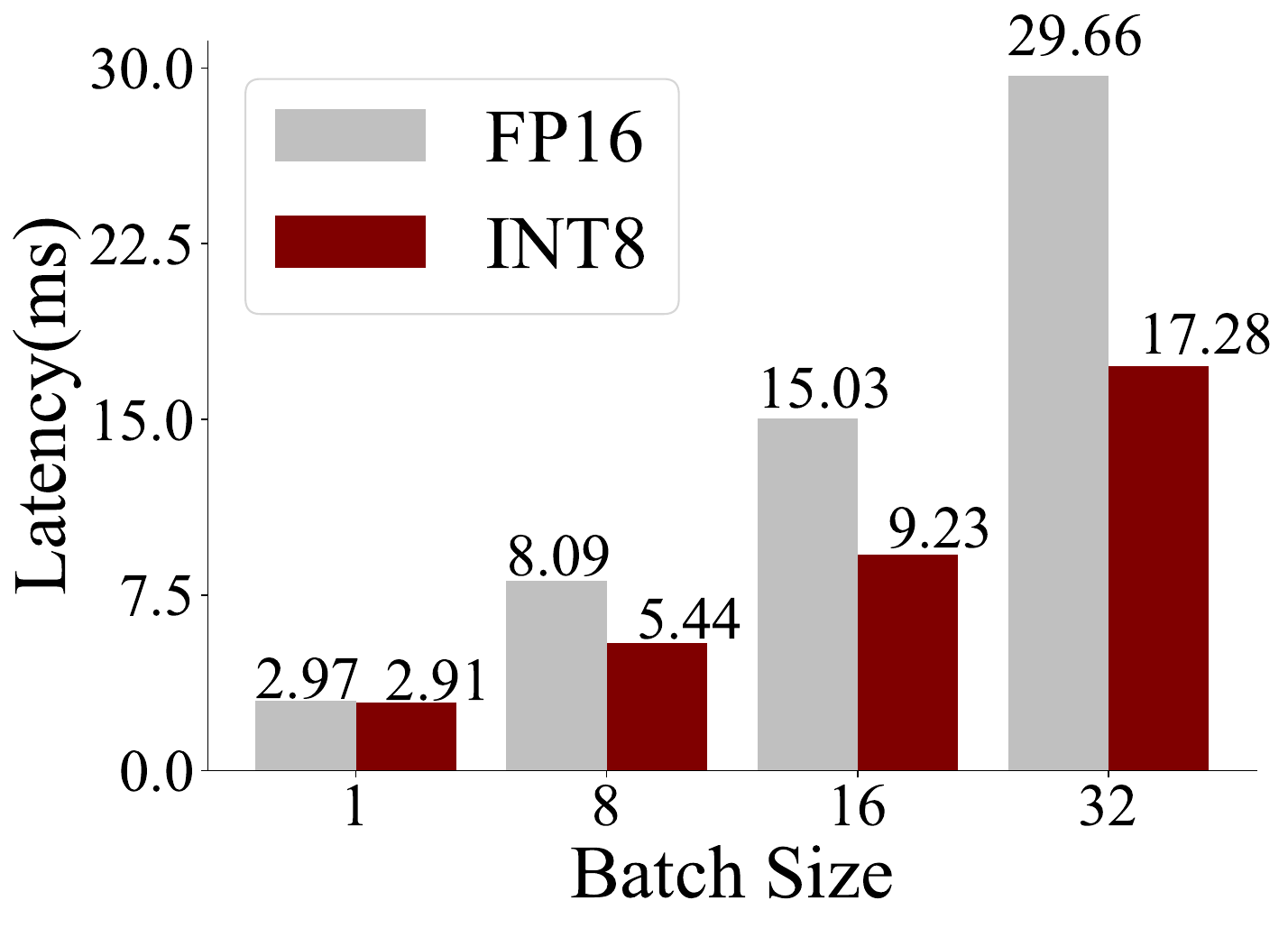}
		\subcaption{ViT-L\_16-224}
	\end{minipage}
	\begin{minipage}[t]{0.25\linewidth}
		\centering
		\includegraphics[width=0.9\linewidth]{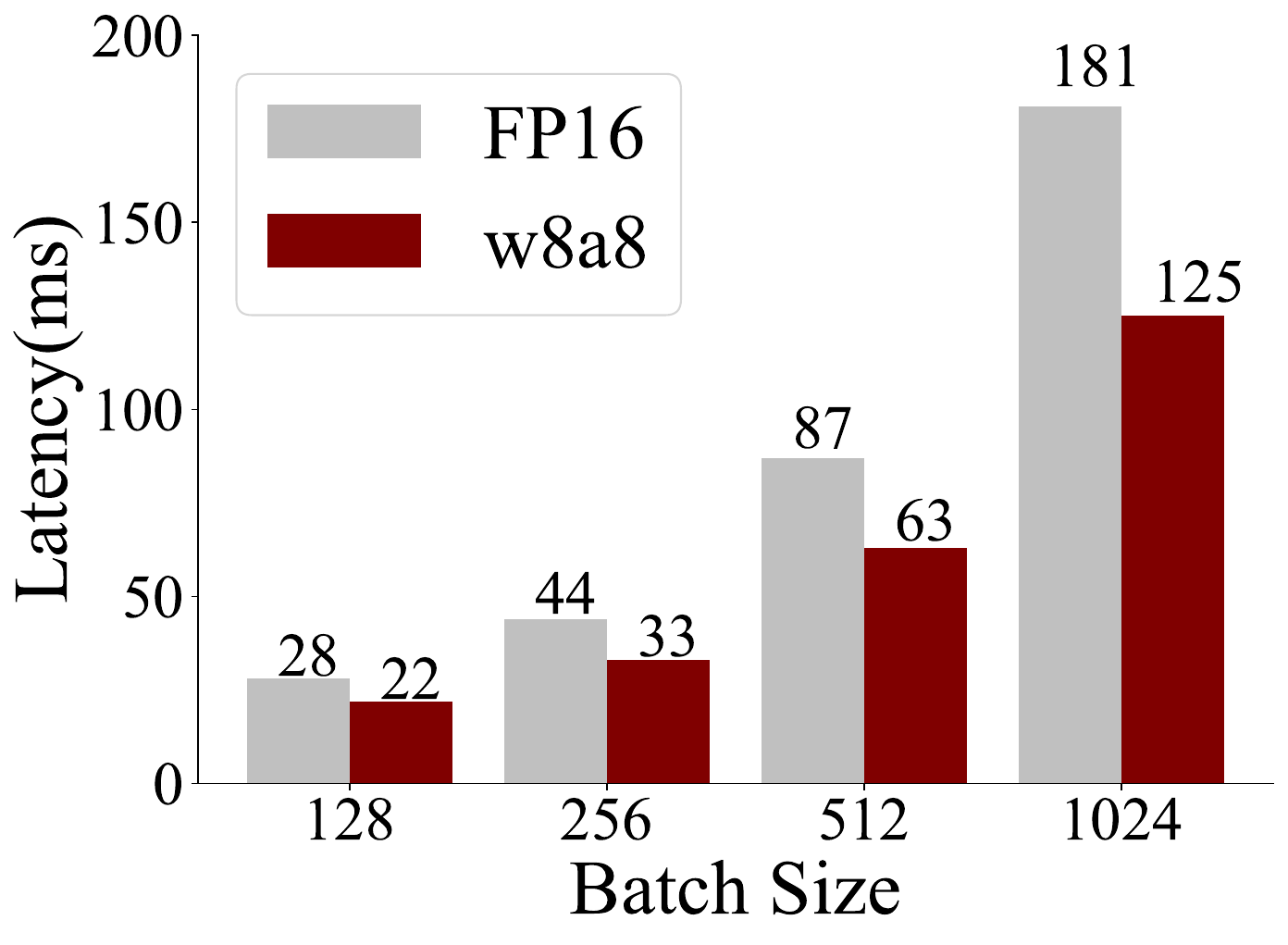}
		\subcaption{OPT-13B}
	\end{minipage}%
	\begin{minipage}[t]{0.25\linewidth}
		\centering
		\includegraphics[width=0.9\linewidth]{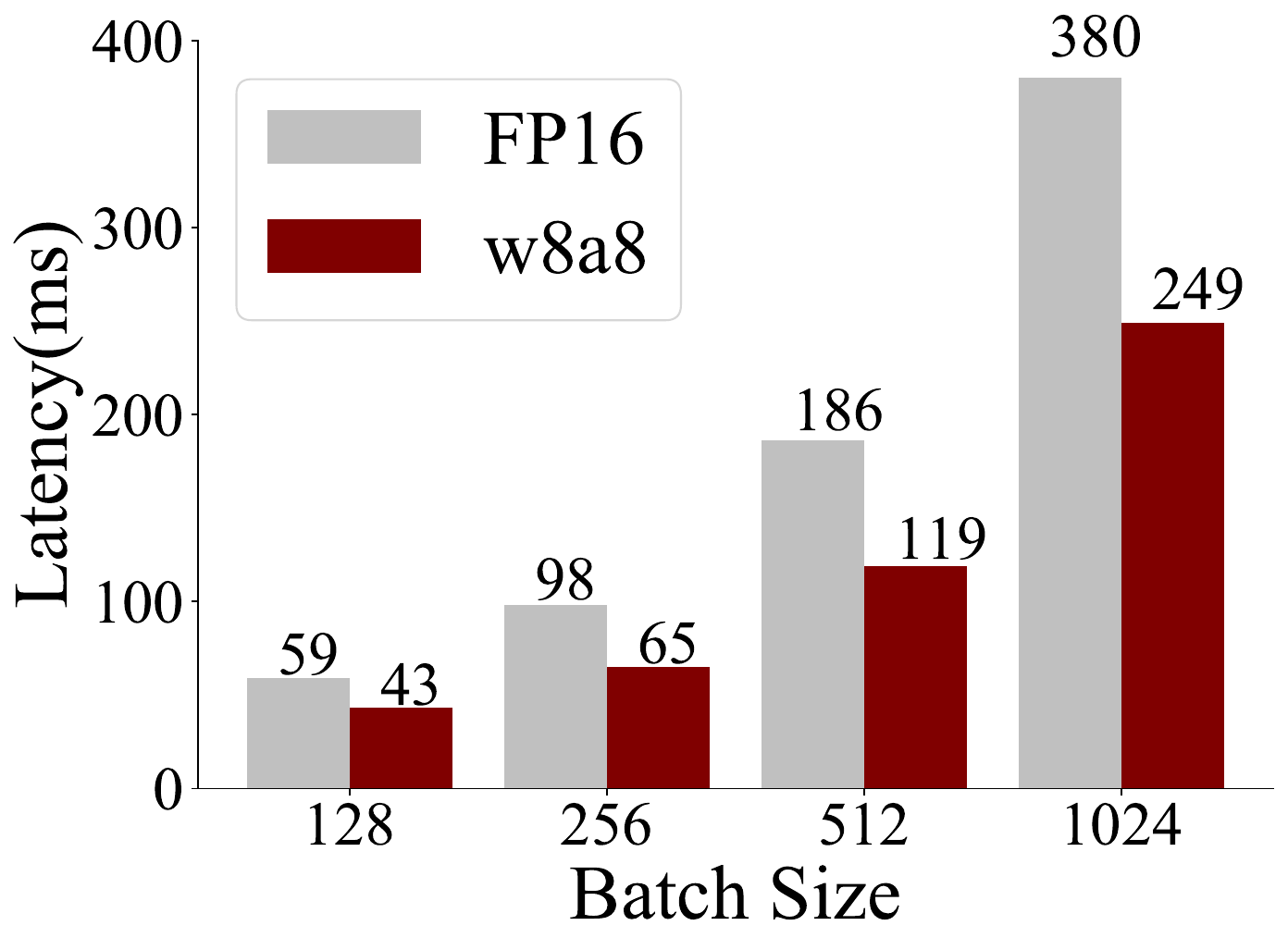}
		\subcaption{OPT-30B}
	\end{minipage}
	\begin{minipage}[t]{0.25\linewidth}
		\centering
		\includegraphics[width=0.9\linewidth]{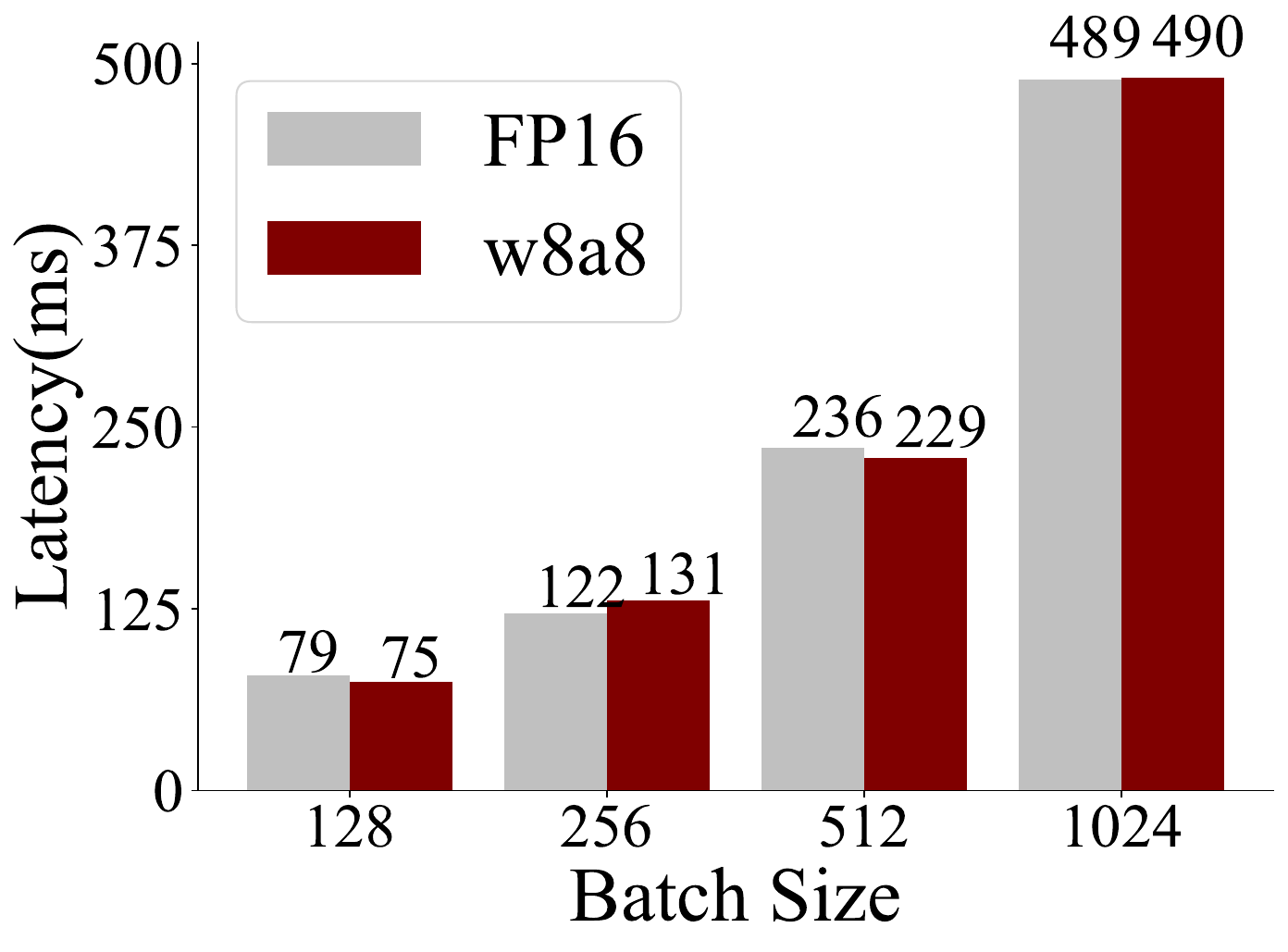}
		\subcaption{OPT-66B}
	\end{minipage}%
	\begin{minipage}[t]{0.25\linewidth}
		\centering
		\includegraphics[width=0.9\linewidth]{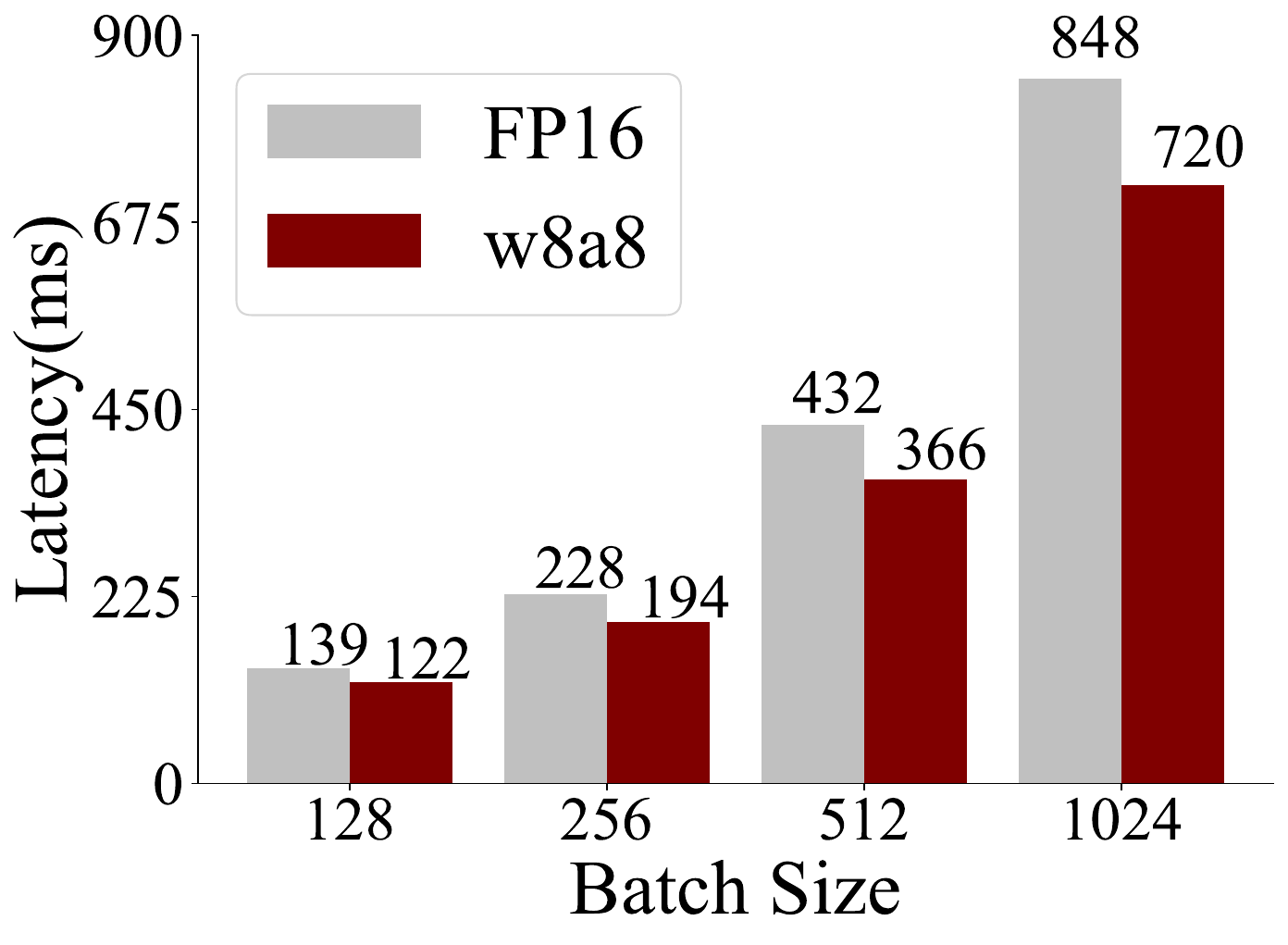}
		\subcaption{OPT-175B}
	\end{minipage}
	\vspace{-2mm}
	\caption{Inference latency of the ViT and OPT using FasterTransformer on NVIDIA A100-80GB GPUs. The data of OPT is from~\cite{xiao2023smoothquant}.}
	\label{latency_ViT_OPT}
	\vspace{-4mm}
\end{figure*}

\subsubsection{Quantization for Transformer-Based Large Language Models}
Before 2023, the study of quantization~\cite{shen2020q, bondarenko2021understanding, bai2022towards, zafrir2019q8bert, tang2022mkq, kim2021bert, bai2020binarybert, qin2022bibert, zhang2020ternarybert, zhao2021automatic} on Transformer-based NLP focused almost entirely on the BERT architecture. With the popularity of pretrained large language models~\cite{ouyang2022training, touvron2023llama, gpt4}, researchers~\cite{dettmers2022llm,park2022nuqmm,yao2022zeroquant} stated to explore how to quantize the  Transformer with billions of parameters and explores more efficient quantization schemes with limited data and computational overhead.

\noindent \textbf{Post-training quantization. }
Based on the analysis of outliers for quantized Transformers,  Outlier Suppression~\cite{wei2022outlier} proposes to migrate the gamma of LayerNorm to the next module for that the gamma amplifies the outliers in the output and causes large quantization error, and then clips the tokens in a coarse-to-fine procedure. MREM~\cite{bai2022towards} focuses on reducing the computational cost of quantization, and minimizes the output quantization error for all modules in parallel manner. Zeroquant~\cite{yao2022zeroquant} finds that the performance degradation is due to the different dynamic range of tokens and weights, proposes to adopt group-wise quantization for weights and token-wise quantization for activations, and utilizes knowledge distillation layer by layer. For large Transformers with billions of parameters, the outlier is still the main reason for large accuracy degradation for quantized models.
To address this, LLM.int8()~\cite{dettmers2022llm} represents the activations and outliers of weights with 16-bit and performs 8-bit vector-wise quantization for weight tensor, but the acceleration is limited and even worsened due to the irregular quantization granularity~\cite{xiao2023smoothquant}. GPTQ~\cite{frantar2022gptq} also only quantizes weight parameters as LLM.int8(), but adopts unified quantization strategy. GPTQ quantizes the weights based on approximate second-order information, thereby getting much more accurate quantized weight in a few hours. AWQ~\cite{lin2023awq} proposes to search for the optimal per-channel scale factors by observing the distribution of activations instead of weights, allowing the LLMs retain the capabilities for different domains and modalities. Outlier Suppression+~\cite{wei2023outlier} explores more accurate outliers suppression schemes with channel-wise shifting and scale, thereby helping align the range of different activation channels and scale down the outliers. The shifting and scale factors could also be merged with other weight parameters. Similarly, SmoothQuant~\cite{xiao2023smoothquant} and QLLM~\cite{liu2023qllm} propose the mathematically equivalent per-channel scaling transformation that migrates the quantization difficulty from activations to weights. To further improve the performance of quantized LLMs, QLLM also learns low-rank parameters by minimizing the reconstruction error of the outputs between floating-point and quantized LLMs with limited calibration data. Similarly, RPTQ~\cite{yuan2023rptq} utilizes a reorder-based scheme that rearranges the channels of activations with a similar range and quantizes them in cluster, and then migrates the scale into LayerNorm and weights of linear layers without extra computational overhead in the inference.Based on scale migration, OmniQuant~\cite{shao2023omniquant} further proposes a learnable PTQ method module by module, where the weight clipping parameters and transformation scale of activations are optimized with gradient descent. 
To minimize error accumulation in adjacent blocks, CBQ~\cite{ding2023cbq} presents a cross-block reconstruction framework, 
simultaneously learn the rounding matrices of weight and step sizes of weights and activations. The rounding matrices are learned with LoRA technique, which does not bring much extra cost for PTQ. 
Like GPTQ, SqueezeLLM~\cite{kim2023squeezellm} also focuses on quantizing weights for the memory bandwidth and proposes to search for the optimal bit based on second-order information. Also, SqueezeLLM does not suppress the outliers and sensitive weight values but stores them in an efficient sparse format for more accurate quantized LLMs. Unlike the previous methods, SignRound~\cite{cheng2023optimize} proposes to optimize quantized LLMs from the perspective of adaptive rounding. SignRound designs block-wise tuning using signed gradient descent to learn the weight rounding, thereby greatly helping the output reconstruction of each block.
\begin{table}[t]
		\centering
		\caption{Perplexity (PPL) comparison of different quantization methods on WikiText2. W/E denotes weights and embedding, respectively. * represents adopting different methods to get the perplexity, please refer to the corresponding papers.}
		\label{Quant4LLM}
		\vspace{-2mm}
		\setlength{\tabcolsep}{3pt}
		\scalebox{1.0}{
			\begin{tabular}{l  c c |c c c c}
				\toprule[0.1em]
				\multirow{2}{*}{\textbf{Method}} & \multirow{2}{*}{\textbf{PTQ}} & {\bf Bits} & \multicolumn{4}{c}{\textbf{LLaMA}}\\
				& & (W-E)  & 7B & 13B & 30B & 65B \\
				\midrule
				\midrule
				FP16 & $\times$ & 16/16 & 5.68 & 5.09 & 4.10 & 3.56   \\
				\midrule
				SmoothQuant*~\cite{xiao2023smoothquant} & \checkmark & 8/8  & 11.56 & 10.08 & 7.56 & 6.20 \\
				LLM-QAT*~\cite{liu2023llm} & $\times$ & 8/8 & 10.30 & 9.50 & 7.10 & -\\
				\midrule
				OS+~\cite{wei2023outlier} & \checkmark & 6/6 & 5.76 & 5.22 & 4.30 & 3.65   \\
				OmniQuant~\cite{shao2023omniquant} & \checkmark & 6/6  & 5.96 & 5.28 & 4.38 & 3.75 \\
				QLLM~\cite{liu2023qllm} & \checkmark & 6/6 & 5.89 & 5.28 & 4.30 & 3.73 \\
				\midrule
				SqueezeLLM~\cite{kim2023squeezellm} & \checkmark & 4/16 & 5.79 & 5.18 & 4.22 & 3.76 \\
				SignRound~\cite{cheng2023optimize} & \checkmark & 4/16 & 6.12 & 5.32 & 4.52 & 3.90 \\ 
				OmniQuant~\cite{shao2023omniquant} & \checkmark & 4/16  & 5.86 & 5.21 & 4.25 & 3.71 \\
				\midrule
				LLM-QAT*~\cite{liu2023llm} & $\times$ & 4/8  & 10.90 & 10.00 & 7.50 & -\\
				\midrule
				OS+~\cite{wei2023outlier} & \checkmark & 4/4- & 14.17 & 18.95 & 22.61 & 9.33   \\
				OmniQuant~\cite{shao2023omniquant} & \checkmark & 4/4  & 11.26 & 10.87 & 10.33 & 9.17 \\
				QLLM~\cite{liu2023qllm} & \checkmark & 4/4 & 9.65 & 8.41 & 8.37 & 6.87 \\
				PEQA~\cite{kim2023memory} & $\times$ & 4/4 & 5.84 & 5.30 & 4.36 & 4.02  \\
				\midrule
				SqueezeLLM~\cite{kim2023squeezellm} & \checkmark & 3/16 & 6.32 & 5.60 & 4.66 & 4.05 \\
				OmniQuant~\cite{shao2023omniquant} & \checkmark & 3/16 & 6.49 & 5.68 & 4.74 & 4.04 \\
				\midrule
				PEQA~\cite{kim2023memory} & $\times$ & 3/3 & 6.19 & 5.54 & 4.58 & 4.27  \\
				\midrule
				OmniQuant~\cite{shao2023omniquant} & \checkmark & 2/16 & 15.47 & 13.21 & 8.71 & 7.58 \\
				\bottomrule[0.1em]
		\end{tabular}}
		\vspace{-2mm}
	\end{table}

\noindent \textbf{Quantization-aware training.} 
Q-BERT~\cite{shen2020q} is the early work that conducts quantization-aware training on Transformer-based architecture for natural language processing. Inspired by HAWQ~\cite{dong2019hawq}, Q-BERT searches mixed-quantization settings based on the second order Hessian information and adopts group-wise quantization to partition parameter matrix into multiple groups to reduce accuracy degradation. I-BERT~\cite{kim2021bert} not only quantizes the linear and self-attention layer, but also designs an integer-only inference scheme for the nonlinear operations (GELU, SoftMax and LayerNorm) in Transformers, which also inspires the FQ-ViT and I-ViT for the quantization of vision Transformer. Specifically, I-BERT proposes to use the second-order polynomial approximate for GELU and exponential function of SoftMax, and calculate the standard deviation of LayerNorm based on Newton’s Method. In this way, I-BERT can achieve faster inference compared to the baseline and normal quantization methods. To compensate for the disadvantages of the hand-crafted heuristics in Q-BERT, AQ-BERT~\cite{zhao2021automatic} proposes an automatic mixed-precision quantization scheme to learn the bit-width and parameters of each layer simultaneously, which is inspired from differentiable network architecture search. AQ-BERT achieves better results than Q-BERT and more suitable for resource-limited devices. Besides the QAT schemes for BERT, there are only a few papers that conducts quantization-aware training on large language models for the huge training overhead. PEQA~\cite{kim2023memory} and QLoRA~\cite{dettmers2023qlora}
explore parameter-efficient fine-tuning techniques to train the quantized LLMs, with the former only optimizing the scale factors while freezing the quantized weights, and the latter adopting 4-bit NormalFloat, double quantization, and paged optimizers for normally distributed weights and reduce memory footprint during training. LLM-QAT~\cite{liu2023llm} proposes a data-free quantization-aware training method, where the training data is generated from the original pretrained large language model. LLM-QAT not only quantizes all the linear layers and self-attention, but also quantizes the KV cache with cross-entropy based logits distillation.

\subsubsection{Quantization for Transformer-Based Vision Models}

\textbf{Post-training quantization.} PTQ-ViT~\cite{liu2021post} firstly explores post-training quantization on vision Transformers,  using the nuclear norm of the attention map and output feature in the Transformer layer to determine the bit-width of each layer. To get more accurate quantized Transformers, PTQ-ViT further proposes a rank loss to keep the relative order of the self-attention of quantized Transformers. 
PTQ4ViT~\cite{yuan2022ptq4vit} observers that the distributions of post-softmax and post-GELU activations are hard to quantize with conventional methods, and so introduces a twin uniform quantizer and use a Hessian guild metric to find the optimal scale parameters. 
To construct a full-quantized vision Transformer, FQ-ViT~\cite{lin2021fq}  not only quantizes all the linear and self-attention layers, but also quantizes LayerNorm and SoftMax with the power-of-two factor and Log2 quantizer.
APQ-ViT~\cite{ding2022towards} explores block-wise optimization scheme to determine the optimal quantizer for extremely low-bit PTQ, and utilizes an asymmetric linear quantization to quantize the attention map for maintaining the Matthew-effect of Softmax.
Based on the observation that the normal uniform quantizer can not effectively handle the heavy-tailed distribution of vision Transformer activations, NoisyQuant~\cite{liu2023noisyquant} proposes to enhance the quantizer by adding a fixed uniform noise following the theoretical results, refining the quantized distribution to reduce quantization error with minimal computation cost.
For the post-LayerNorm and post-SoftMax activations that have extreme distributions, RepQ-ViT~\cite{li2023repq} applies channel-wise quantization to deal with the severe inter-channel variation of the former and utilize log$\sqrt{2}$ quantizer to compress the power-law features of the later one. Before the inference, RepQ-ViT reparameterizes the scale factors to the layer-wise and log2 quantizer with mere computations.

\begin{figure*}[h]
	\centering
	\includegraphics[width=0.95\linewidth]{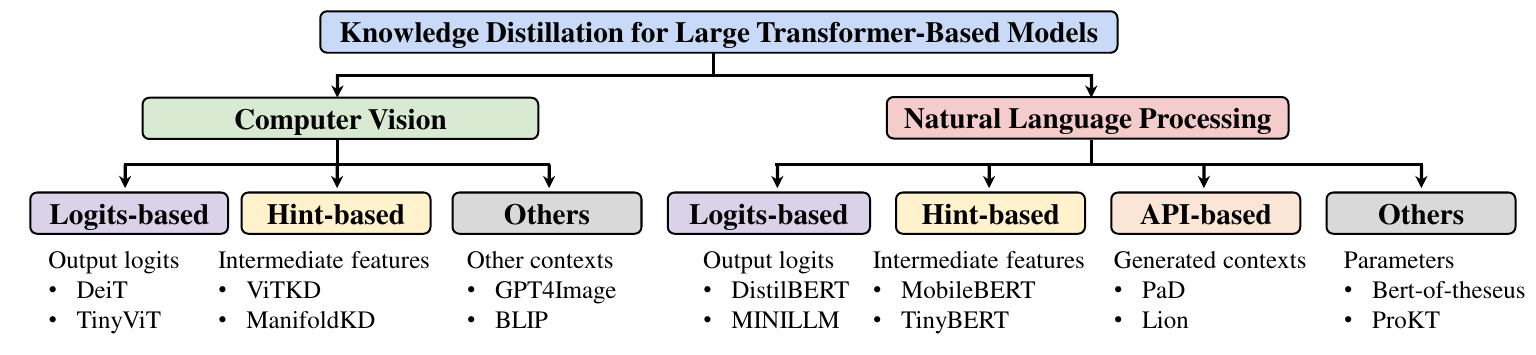}
	\vspace{-0.2cm}
	\caption{The taxonomy of knowledge distillation used for large Transformer-based models.}
	\label{fig:kd_taxonomy}
\end{figure*}

\noindent \textbf{Quantization-aware training.}  When compressing vision Transformers to extremely low-bit precision, PTQ can not optimize the large quantization error with limited calibration images and suffers from significant performance reduction. As such, QAT is urgently required for more accurate low-bit vision Transformers. 
Q-ViT~\cite{li2022q_diff} finds that MSA and GELU are highly sensitive to quantization, and so proposes a fully differentiable quantization method that adopts head-wise bit-width and switch scale during the quantization searching process.
Quantformer~\cite{wang2022quantformer} takes the self-attention rank into consideration, and proposes to maintain the consistency between quantized and full-precision vision Transformers. In addition, Quantformer presents the group-wise strategy to quantize feature of patches in different dimensions, where each group adopts different quantization parameters and the extra computation cost is negligible.
Based on the observation that sever performance degradation suffers from the quantized attention map, AFQ-ViT~\cite{li2022q} designs an information rectification module and a distribution guided distillation during quantization training. The former helps recover the distribution of attention maps with information entropy maximization in the inference, and the latter reduces the distribution variation with attention similarity loss in the backward. Similar with FQ-ViT~\cite{lin2021fq}, I-ViT~\cite{li2023vit} also explores the integer-only quantization scheme for ViTs. I-ViT designs Shiftmax, ShiftGELU and I-LayerNorm to replace the vanilla modules with bit-wise shift and integer matrix operations in the inference, achieving $3.72-4.11\times$ acceleration compared to the floating-point model. OFQ~\cite{liu2023oscillation} finds that weight oscillation causes unstable quantization-aware training and leads to sub-optimal results, and the oscillation comes from the learnable scale factor and quantized query and key in self-attention. To address that, OFQ proposes statistical weight quantization to improve quantization robustness, freezing the weights with high confidence and calming the oscillating weights with confidence-guided annealing. For the query and key in the self-attention, OFQ presents query-key reparameterization to decouple the negative mutual-influence between quantized query and key weights oscillation. Similarly, VVTQ~\cite{huang2023variation} analyzes ViT quantization from the perspective of variation, which indicates that the data variance within mini-batch is much harmful to quantization and slow down the training convergence. To reduce the impact of variations in the quantization-aware training, VVTQ proposes a multi-crop knowledge distillation-based quantization methodology, and introduces module-dependent quantization and oscillation-aware regularization to enhance the optimization process.

We summarize the results of these PTQ and QAT methods in Table~\ref{Quant4CV}. Most schemes conduct PTQ on 8-bit and 6-bit, and there is server degradation when quantized to 4-bit. With the complete training data, QAT could push the ViT quantization to 4-bit and less. Some methods even show better results than floating point models, which prove that quantization-aware training could tap into the true accuracy of quantized models efficiently. Figure~\ref{latency_ViT_OPT} shows the latency results of ViT models using FasterTransformer\footnote{https://github.com/NVIDIA/FasterTransformer}.

\subsubsection{Discussion}
With the presence of extreme distributions and outliers, quantization for Transformers is much more difficult than that for convolutional neural networks. To recover the performance of quantized models, various methods are proposed to address the quantization-unfriendly components in Transformers. Specifically, for vision tasks, the existing methods optimizes the quantized Transformers in three ways: retaining the self-attention rank, rectifying the distribution of extreme distribution, and addressing the weight oscillation and data variation in quantization-aware training, as shown in Figure~\ref{quant_overview}. And for natural language processing task, most schemes aims to process the outlier of weights and activations. What's more, the training overhead is usually unacceptable for large language models with billions of parameters, and therefore quantizing module by module and conducting parameter-efficient fine-tune are much popular. However, when compressing to extremely low bit-width, the quantized Transformers suffer significant performance degradation, and perform far worse than the floating-point models. As such, how to build a more accurate and low-bit Transformers is still a difficult problem to be solved.

\subsection{Knowledge Distillation}

In this section, we will introduce knowledge distillation frameworks used in compressing Transformer based foundation models, including both language and vision models.

\subsubsection{Overview of Knowledge Distillation}

Knowledge distillation (KD) aims to train student networks by compressing~\cite{ba2014deep, bucilua2006model,wu2023weight} or transferring~\cite{hinton2015distilling} knowledge from teacher networks. In this paper, our main focus lies on distillation methods proposed to achieve a compact student model while preserving satisfactory performance compared to a cumbersome teacher model. The student models typically have narrower and shallower architectures, making them more suitable for deployment on resource-limited systems.

We will first discuss two different kinds of knowledge distillation in the following: logits-based methods~\cite{hinton2015distilling,wu2024rethinking,tang2019distilling,ko2024distillm} which convey knowledge on the logits level and hint-based methods~\cite{romero2014fitnets} which convey knowledge through intermediate features. To illustrate the objective of logits-based KD in classification tasks, we denote the logits output of the teacher and student network as $z^{\bf s}, z^{\bf t}\in \mathbb{R}^C$, where $C$ represents the number of classes. Neural networks often generate class probabilities by applying a softmax function to the logits, converting them into probabilities $p^{\bf s}, p^{\bf t}$ as follows:
\begin{equation}
	p_i^{\bf s}=\frac{\exp(z_i^{\bf s})}{\sum_{j=1}^{C}\exp(z_j^{\bf s})},\quad p_i^{\bf t}=\frac{\exp(z_i^{\bf t})}{\sum_{j=1}^{C}\exp(z_i^{\bf t})},
\end{equation}
With above probabilities, the logits-based KD minimizes the KL divergence between the probabilities of the teacher and student model as:
\begin{equation}
	L_{logits}=KL(p^{\bf t}||p^{\bf s})=\sum_{j=1}^{C}p_j^{\bf t}log(\frac{p_j^{\bf t}}{p_j^{\bf s}}),
	\label{eq:KLD}
\end{equation}
As for hint-based KD, given the intermediate features of the teacher and student network, $F^{\bf s}, F^{\bf t}\in \mathbb{R}^{H\times W\times C}$, the corresponding loss is formulated as:
\begin{equation}
	L_{hint}=\mathcal{H}(F^{\bf s}, F^{\bf t})=||F^{\bf t}-\phi(F^{\bf s})||^2,
\end{equation}
where $\phi$ is a function used to ensure that the student features have the same shape as the teacher features. $\mathcal{H}$ represents the chosen metric function, here we provide an example using mean squared error.

In addition to the two mainstream KD methods mentioned above, we will also discuss an API-based method, where only the teacher's generated outputs are accessible in today's large language models (LLMs).

\subsubsection{KD for Transformer-Based Large Language Models}
\noindent\textbf{Logits-based KD.}
Tang~\etal~\cite{tang2019distilling} employs knowledge distillation to compress BERT\cite{devlin2018bert}, a large language model, into a much lighter bidirectional long short-term memory network (BiLSTM)~\cite{hochreiter1997long} for natural language processing (NLP) tasks. The distillation process aims to minimize the mean squared error (MSE) loss between the student's logits and the teacher's logits. Following distillation, the shallow BiLSTM-based model achieves comparable results to ELMo\cite{peters1802deep}, but with approximately 99\% fewer parameters and a 15$\times$ faster inference speed. Similarly, DistillBERT~\cite{sanh2019distilbert} initializes the shallower student with teacher’s parameters, and minimized the soft target probabilities between the teacher and the student, also known as word-level KD.
SeqKD~\cite{kim2016sequence} fine-tunes the student model on the sequence-level teacher-generated data.
MixKD~\cite{liang2020mixkd} extends the idea of encouraging the student to mimic the teacher's logits to the linear interpolation of example pairs~\cite{zhang2017mixup}. Theoretical analysis has shown that, under reasonable conditions, MixKD can effectively reduce the gap between generalization error and empirical error.
Turc~\etal~\cite{turc2019well} demonstrates the ongoing significance of pre-training, even when employing smaller architectures. They introduce a methodology called Pre-trained Distillation (PD), which begins with pre-trained compact student models. Subsequently, it explores the transfer of task knowledge from large fine-tuned models using conventional logits-based KD.
In contrast, MINILLM~\cite{gu2023knowledge} highlights a limitation of conventional logits-based KD that minimize forward Kullback-Leibler divergence (KLD) in free-run generation. They find that such approaches can lead the student model to overestimate the low-probability regions of the teacher's distribution. To address this issue, MINILLM proposes replacing the forward KLD objective (as shown in Eq.~\ref{eq:KLD}) with reverse KLD, \ie, $reverse~\mathrm{KLD}:=KL(p^{\bf s}||p^{\bf t})$, which is better suited for knowledge distillation on generative large language models (LLMs).
GKD~\cite{agarwal2023gkd} identifies a challenge in distilling knowledge into auto-regressive student models, which is the train-test distribution mismatch. Specifically, partial sequences encountered by the student during generation phase can be significantly different from the ones observed during the training phase. To address this issue, GKD trains the student on its self-generated output sequences by leveraging feedback (logits) from the teacher on such sequences. It also facilitates the integration of distillation with RL fine-tuning of LLMs. Huang~\etal~\cite{huang2022context} further combines the hint-based KD with multitask in-context learning.

\noindent\textbf{Hint-based KD.}
Sun~\etal~\cite{sun2019patient} proposes patient knowledge distillation to help shallow BERT student learn from multiple intermediate features~\cite{bae2023fast,sun2020mobilebert,jiao2019tinybert,wang2020minilm} of the deep BERT teacher.
Li~\etal~\cite{li2019hint} utilizes hints extracted from both intermediate hidden states and attention distributions to enhance the training of Non-AutoRegressive Translation (NART) models. The distilled NART models attain performance similar to a powerful LSTM-based AutoRegressive Translation (ART) baseline in various machine translation tasks. Notably, the distilled NART models exhibit a speed improvement of 17$\times$ over their ART counterparts. Similarly, Mukherjee~\etal~\cite{mukherjee2020xtremedistil} employs MBERT~\cite{tsai2019small} as a teacher model to guide the training of smaller BiLSTM models, resulting in a significant 51$\times$ speed improvement for batch inference. 
MobileBERT~\cite{sun2020mobilebert} implements two objective functions to distill knowledge from a BERT teacher incorporated with inverted bottleneck, including attention distributions and hidden states, to a slimmed-down version of BERT as the student model. In addition to these objectives, TinyBERT~\cite{jiao2019tinybert} extends the distillation process by also transferring knowledge from the teacher's logits to the student model.
Wang~\etal~\cite{wang2020minilm} proposes MINILM to train the student networks to mimic both the attention distributions and the scaled dot-product between values of teacher models. It retains more than 99\% accuracy on General Language Understanding Evaluation (GLUE) benchmark tasks using 50\% of the Transformer parameters and computations of the teacher model.

\noindent\textbf{API-based KD.}
Today's language models have reached unprecedented scale~\cite{brown2020language,wei2022emergent, schaeffer2023emergent} in terms of computation and model parameters, as evident from the significant advancements made in large language models (LLMs) like GPT-4~\cite{gpt4}. However, the power of such models is only accessible through their APIs, which limits the ability to utilize conventional KD methods that rely on intermediate features or logits of the teacher model. In this section, we will discuss API-based KD, also known as black-box KD, where only the teacher's final generations are accessible for the purpose of distillation.

Several recent studies~\cite{ho2022large,hsieh2023distilling,magister2022teaching,wang2023scott,saha2023can,jiang2023lion,wu2023lamini,li2022explanations,west2021symbolic} have showed the promising results in fine-tuning small models on the outputs generated by LLMs. Specifically, they first prompt a very large teacher model, such as GPT-3 with 175B parameters, to solve complex questions via chain-of-thought (CoT) reasoning~\cite{wei2022chain}. Then, they use the generated instructions~\cite{jiang2023lion,wu2023lamini}, explanations~\cite{li2022explanations} or rationales~\cite{ho2022large} to fine-tune a much smaller student. Furthermore, Zhu~\etal~\cite{zhu2023pad} proposes PaD, which utilizes program-aided reasoning, such as an additional Python interpreter, to help student models overcome faulty steps in the CoT reasoning with automated error checking.
Shridhar~\etal~\cite{shridhar2023distilling} introduces Socratic CoT that trains a combination of two small distilled models: a problem decomposer and a subproblem solver. These models work together to decompose and solve complex problems effectively.
Fu~\etal~\cite{fu2023specializing} demonstrates that specializing the small student model's abilities for the specific reasoning task, by transferring the knowledge in larger teacher model's generic directions, yields promising results.
Jiang~\etal~\cite{jiang2023lion} identifies a shortcoming in the aforementioned approaches, as they lack the ability to incorporate feedback for identifying challenging instructions where the student model's performance falls short. To address this limitation, they propose a three-stage adversarial loop of distillation that incorporates feedback and addresses these challenging instructions.

\noindent\textbf{Other KDs.}
Xu~\etal~\cite{xu2020bert} compresses the BERT model by progressive module replacing. The original model is first divided into multiple modules, and more compact substitutes are created for these modules. Then, they randomly substitute the original modules with their corresponding substitutes and train the compact modules without any additional loss functions.

Jha~\etal~\cite{jha2023large} discovers that fitting both the student and teacher models, each with numerous parameters, into GPU memory with traditional knowledge distillation is impractical. To overcome this limitation, they propose a teacher-free task-agnostic distillation method which uses a truncated version (student) of the large model (teacher) for initialization~\cite{li2021short} and then continuing the pre-training of student, without employing any distillation loss.

\begin{table*}[t]
	\centering
	\small
	\renewcommand\arraystretch{1.0}
	\setlength{\tabcolsep}{6pt}
	\caption{\small{Comparison with previous Transformer-based language model distillation approaches. The GLUE score is averaged on 8 tasks, \ie, SQuAD2, MNLI-m, SST-2, QNLI, CoLA, RTE, MRPC, QQP.}}
	\vspace{-0.5em}
	\begin{tabular}{l|c|c|c|c|c|c}
		\Xhline{1pt}
		Models & Distillation & Teacher & $\#$ Layer & $\#$ Params & Speed Up & GLUE \\		
		\hline
		BERT$_{BASE}$~\cite{devlin2018bert}  & - & - & 12 & 110M & $\times$1 & 79.6$^\dagger$ (81.5$^\ddagger$) \\
		BERT$_{LARGE}$~\cite{devlin2018bert} & - & - & 24 & 340M & - & 81.9$^\dagger$ \\
		\hline
		DistilBERT~\cite{sanh2019distilbert} & logits & BERT$_{BASE}$ & 4 / 6 & 52M / 66M & $\times$3.0 / $\times$2.0 & 71.2 / 76.2 (75.2$^\ddagger$) \\
		TinyBERT~\cite{jiao2019tinybert} & hint + logits & BERT$_{BASE}$ & 4 / 6 & 15M / 66M & $\times$9.4 / $\times$2.0 & 76.5 / 79.0 (79.1$^\ddagger$) \\
		BERT-PKD~\cite{sun2019patient} & hint + logits & BERT$_{BASE}$ & 3 / 6 & 46M / 66M & $\times$3.7 / $\times$1.6 & 76.0$^\dagger$ / 80.6$^\dagger$ \\
		MobileBERT~\cite{sun2020mobilebert} & hint & BERT$_{LARGE}$ & 24 & 25M & $\times$4.0 & 79.7$^\dagger$ \\
		PD~\cite{turc2019well} & logits & BERT$_{BASE}$ & 6 & 66M & $\times$2.0 & 81.2$^\dagger$ \\
		MINILM~\cite{gu2023knowledge} & hint & BERT$_{BASE}$ & 6 & 66M & $\times$2.0 & 80.4$^\ddagger$ \\
		\Xhline{1pt}
	\end{tabular}
	\label{table:kd_bert}
	\begin{tablenotes}
		\item $^\dagger$ The corresponding data is from~\cite{qiu2020pre}.
		\item $^\ddagger$ The corresponding data is from~\cite{gu2023knowledge}. They reported F1 for SQuAD 2.0, and accuracy for other datasets.
	\end{tablenotes}
\end{table*}

\subsubsection{KD for Transformer-Based Vision Models}
\noindent\textbf{Logits-based KD.}
While conventional KD in vision tasks typically involves using soft labels~\cite{hinton2015distilling,yu2022unified,hao2023vanillakd} generated by a powerful teacher network, Touvron~\etal~\cite{touvron2021training} makes a noteworthy discovery. Their experiments on the ImageNet benchmark reveal that employing hard labels, which are essentially the maximum of the score values predicted by the teacher model, leads to superior results. Additionally, they introduce a novel distillation token into vision Transformers (ViTs)\cite{dosovitskiy2020image}.
TinyViT~\cite{wu2022tinyvit} observes that smaller ViTs can benefit from larger teacher using massive pre-training data, for example, pre-training (distilling) student on ImageNet-21k while fine-tuning student on ImageNet-1k. And they propose TinyViT to save computational memory by storing data augmentation information and logits in advance for large teacher models.
Ren~\etal~\cite{ren2022co} argues that student accuracy is primarily influenced by the inductive bias of the teacher models rather than their accuracy. To address this, they introduce a cross-inductive bias distillation approach, which involves distilling the student model with multiple teachers exhibiting distinct architectural inductive biases.
To enhance the utilization of the inductive bias in CNNs and offer more image-friendly guidance, CSKD~\cite{zhao2023cumulative} removes the pooling operation after the final feature map and treats the features at each position as individual samples. Subsequently, these position-based features are fed into the classifier, leading to the generation of dense predictions for each corresponding position. The hard labels derived from the CNN's dense predictions are used as target labels, and cross-entropy serves as the loss function for spatial knowledge transfer.

\noindent\textbf{Hint-based KD.}
Chen~\etal~\cite{chen2022dearkd} introduces a two-stage DearKD framework. In the initial stage, knowledge distillation is employed from the intermediate layers of the CNN to help Transformer-based student capture the inductive biases. Subsequently, the second stage trains the student without distillation.
Hao~\etal~\cite{hao2022learning} proposes a fine-grained manifold distillation method to fully utilize the patch-level and batch-level information inside Transformer-based architectures. Specifically, they manually select intermediate layers from teacher-student pair and distilled the student via decoupled relation maps.
ViTKD~\etal~\cite{yang2022vitkd} delves into the characteristics of feature maps in ViTs and formulated three practical guidelines for hint-based distillation in ViTs, namely: (i) generating hints is more effective than mimicking in deeper layers, (ii) distillation on shallower layers remains suitable for ViTs distilled through mimicking, and (iii) features from the feed-forward network (FFN) are better than those from the multi-head attention (MHA) for the distillation process.

\subsubsection{Discussion}

Transformer-based language models have shown strong performance on several NLP tasks including text classification, machine translation, speech recognition, question-answering, dialogue systems, \etc. Knowledge distillation is a common and practical technique used to reduce the high computational resource demand of LLMs. Different models, such as T5~\cite{raffel2020exploring}, BERT~\cite{devlin2018bert}, LLAMA~\cite{touvron2023llama}, GPT-3~\cite{brown2020language}, and GPT-4~\cite{gpt4}, exhibit varying performance on different NLP downstream tasks. Therefore, selecting appropriate teachers and methods for model compression is crucial. For example, when compressing the BERT series, we compare various distillation methods in Table~\ref{table:kd_bert}. In contrast to using only logits, hint-based KD methods can transmit richer intermediate layer information to the student, making the learning process easier for the student and yielding better results. However, performing layer-to-layer knowledge distillation sometimes requires carefully designed layer mappings between the teacher and student models. In certain domains where generative models excel, such as reasoning and language understanding, the effectiveness of strong teacher models can only be accessed through their APIs. The exploration of guiding LLMs to generate better outputs for distillation is still in its early stages.

One of the strengths of ViT models is their ability to scale to high parametric complexity, but this requires substantial computational resources and comes at a high cost. KD can be used to transfer knowledge into more compact student models, but there are still challenges in the context of vision that require further research.
The first challenge is related to training costs. Both logits-based and hint-based KD methods require large GPU memory for the distillation process. Vision tasks differ from NLP as the input for vision is in the form of images, which are significantly larger in size compared to the limited sequence length used in NLP tasks. During distillation, even though the teacher network doesn't require backward propagation, each forward pass consumes a significant amount of GPU memory due to the activations of intermediate features. Therefore, finding ways to reduce training costs and shorten training times is an area that needs exploration.
Additionally, choosing the most suitable teacher model for Transformer-based students is an open question. For instance, in the context of ViTs, determining whether a CNN or a Transformer teacher is more effective for the student model is a potential research avenue.
Moreover, as proposed by Hao~\etal~\cite{hao2023vanillakd}, there is a necessity for designing and evaluating KD approaches within practical scenarios, moving away from the limitations of small-scale datasets. This ensures that KD methods are applicable and effective in real-world, large-scale settings.

\section{Architecture Adaptive Compression} 
\label{sec:apt}

\subsection{Pruning}

\subsubsection{Overview of Pruning}
\begin{figure*}[h]
	\centering
	\includegraphics[width=0.95\linewidth]{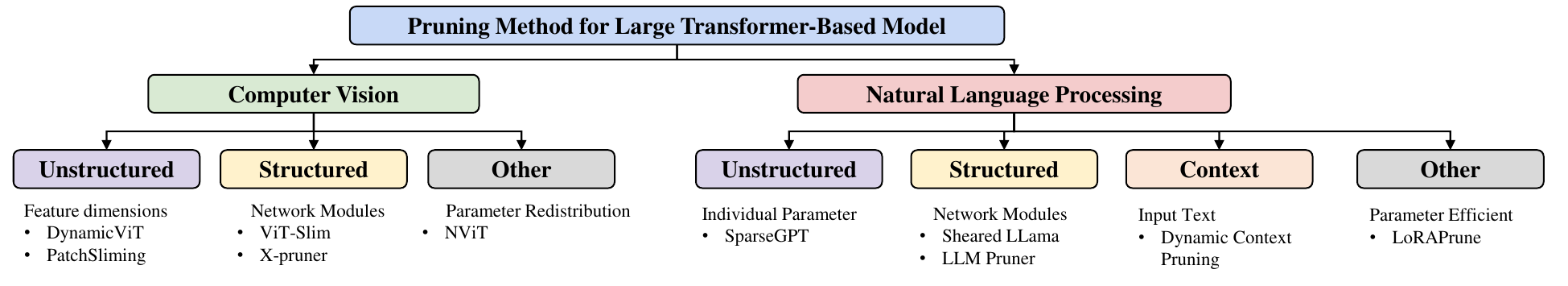}
	\caption{The taxonomy of pruning methods used for Transformer models.}
	\label{prune_overview}
	\vspace{-2mm}
\end{figure*}

Neural network pruning has long been recognized as an effective method for making models more compact and speeding up model inference. The taxonomy of pruning methods can be quite complicated, including the sequential order of pruning and model training, structure specification, and the way to determine the pruned parameters~\cite{Cheng2023ASO}.This study, however, limits the scope of the source model as a pre-trained large Transformer for natural language processing~\cite{Devlin2019BERTPO,brown2020language} and for visual recognition~\cite{Dosovitskiy2020AnII,Touvron2020TrainingDI,Liu2021SwinTH}, raising several specific categories of techniques to be addressed (Fig.~\ref{prune_overview}). 

As pre-training accounts for most of the model performance for downstream tasks, pruning after training (PAT)~\cite{Renda2020ComparingRA} becomes a major choice. From a high-level perspective, the whole process consists of pre-training, pruning, and retraining performance recovery. However, the multi-task nature of pre-training and huge computational in the retraining phase also lead to critical issues to be addressed for pruning large Transformer models, as detailed later.

The methods available for pruning are generally split into two categories according to the structure specification of pruning: unstructured pruning and structured pruning. The former conducts pruning at the finest-grained level~\cite{Lee2018SNIPSN,Zhang2018ASD}, i.e., weight-wise pruning, which follows the optimization problem below,
\begin{equation}
	\mathop{\min}_{\mathbf{\theta}} L(\mathbf{\theta}; D),
	s.t. \|\mathbf{\theta}\|_{0} \leq k
\end{equation}
\noindent where $L$ is the general loss function on the dataset $D$, $\mathbf{\theta}$ denotes the model parameter, and $k$ denotes the targeted non-zero weight number.

Although unstructured pruning generally leads to a reduction in the parameter size or memory usage, it cannot guarantee latency speedups because the resulting model shape tends to be irregular, and such speedups requires specific hardware design. In contrast, specifying the eliminated structure as an entire layer, head, or other network unit results in structural pruning~\cite{You2019GateDG}, which can generally shorten the latency on standard hardware. When the specification is a pattern suitable for speeding-up (e.g., a specific ratio of non-zero mask), the pruning is called semi-structural pruning~\cite{Zhou2021LearningNM}. 

In addition to network parameters, reducing the input size is also an appealing direction to explore. With the emphasis on removing redundant information between tokens, such reduction also achieves lower computation FLOPs with unpruned network parameters, as discovered both in computer vision~\cite{Tang2021PatchSF} and language domain~\cite{Goyal2020PoWERBERTAB} Transformers, . 

\subsubsection{Pruning for Transformer-Based Large Language Models}%
\textbf{Pruning granularity.} In response to the considerable parameter increase in the LLM era, there have been early attempts of pruning arise with the focus on \textbf{unstructured and semi-structured pruning}~\cite{Frantar2023SparseGPTML, Sun2023ASA, Syed2023PruneAT}. SparesGPT, for instance, conducts unstructured and semi-structured (mainly 2:4) pruning starting from OPT-175B~\cite{Zhang2022OPTOP} and BLOOM-176B~\cite{scao2022bloom} and achieve 50-60\% sparsity with moderate perplexity increase and performance decrease on downstream datasets. This work demonstrates the feasibility of low-resource pruning in $>$100 B models with moderate sparsity.

Another line of work mainly focuses on the \textbf{structured pruning} method to achieve a more significant speed-up. Language structural pruning has been conducted at multiple granularities. For instance, Michel~\etal~\cite{Michel2019AreSH} discuss the redundancy of attention head, Fan~\etal~\cite{Fan2019ReducingTD} propose a task-specific extraction of pre-trained sub-networks, Santacroce~\etal~\cite{Santacroce2023WhatMI} focus on FFN layers, and Block Pruning~\cite{lagunas-etal-2021-block} is introduced to prune MHA and FFN parameters separately. In comparison, CoFi~\cite{Xia2022StructuredPL} proposes a more comprehensive coarse- and fine-grained pruning method that learns to generate masks at different granularity, i.e., FFN layers, FFN intermediate dimensions, MHA layers, Attention heads, and hidden dimensions to encourage the optimization in the high sparsity domain. In this way, the pruning of a specific parameter can be determined at different granularity in an end-to-end learnable manner. When pruning the general domain pre-trained BERT model to typical downstream datasets, CoFi achieves more than $10\times$ speed-ups with a limited accuracy gap. Later, this method was extended to a larger decoder-based LLM with a 7B parameter size~\cite{Xia2023ShearedLA}. With a pre-defined model shape, the pruned model shows better instruction tuning performance than models of similar size but trained from scratch. Also, specifying the model configuration with uniform size avoids irregularities in model shape and further increases the inference throughput. In contrast to pre-defining the model shape, Ma~\etal~\cite{Ma2023LLMPrunerOT} propose first to identify groups of coupled structures considering the interdependency of parameters in FFN, MHA, and channel-wise groups and conduct the pruning process by group. Experiments at a relatively low compression rate (20\%) demonstrate the varying effectiveness of these grouping strategies. 

\textbf{Pruning criteria.} To determine the structure to be pruned, various metrics have been explored in the context of LLM pruning. For instance, Frantar~\etal~\cite{Frantar2023SparseGPTML} follows~\cite{Frantar2022OptimalBC} to find an optimal mask that can be used for weight reconstruction and uses Hessian Matrix (i.e., the second order indicator of loss change concerning the change of each parameter) to indicate the parameter importance, which is experimentally shown better than a magnitude criteria~\cite{Zhu2017ToPO} and much faster than previous method AdaPrune~\cite{Hubara2021AcceleratedSN}. In comparison, as noted by LLM Pruner~\cite{Ma2023LLMPrunerOT}, the first-order term should also be considered due to the shift of data used in the pruning process compared to the original language model training. Also, it proposes considering the weight importance at different levels and explores different ways of combining the importance information. Besides, Sun~\etal~\cite{Sun2023ASA} also explores the combination of magnitude and norm of the input activations as the pruning criteria, which lowers the computational cost used for Hessian computation.  

\textbf{Learning to prune.} In addition to determining the importance of parameters using, one can also perform extra training, commonly with regularization on sparsity~\cite{He2017ChannelPF,Voita2019AnalyzingMS}. In particular, the success of LLM as a multi-task language processor promotes the development of a pruning method that preserves this multi-task nature, as opposed to the task-specific trails that are more feasible in the traditional pretrain-finetune paradigm~\cite{Xia2022StructuredPL}. In response to this motivation, recent works such as SIMPLE~\cite{Tao2023StructuredPF}, LLM Pruner~\cite{Ma2023LLMPrunerOT} and Sheared LLama~\cite{Xia2023ShearedLA} incorporate the causal language modeling loss in the pruning objective. However, this “general pruning” strategy raises another issue of data selection in the pruning process, as the domain gap between pre-training and pruning may hinder the importance of estimation/performance recovery. As detailed in~\cite{Xia2023ShearedLA}, a dynamic batch loading strategy is applied to adaptively input the data batch from a proper domain to balance the model performance as a multi-task learner.

\textbf{Computation cost for parameter pruning.} Considering the huge numbers of parameters, post-pruning retraining and parameter importance estimation may both generate a significant computational burden for large language models. In an early attempt, Frantar~\etal~\cite{Frantar2023SparseGPTML} consider a one-shot strategy, i.e., only limited data are used to calibrate the Hessian Matrix in importance estimation, and no retraining was employed. From another perspective, LoRAPrune~\cite{Hu2021LoRALA} shows that introducing parameter-efficient tuning such as Low-Rank Adaption can also decrease the computation cost in retraining while compensating for the performance degradation. 

\textbf{Context and token pruning.} Different from the case in the computer vision domain, language sequence can be very long, and long-range reference or induction also calls for the expansion of input tokens (i.e., context) to prompt better language understanding. Also, since the inference cost of LLM is in quadratic complexity with respect to sequence length, this expansion greatly hinders the efficient inference of language models. To tackle these problems, methods have been proposed since BERT era to prune redundancy in context or context attention calculation. For context pruning, Kim~\etal~\cite{Kim2021LearnedTP} propose to prune tokens across the Transformer layers progressively and improves the throughput by several folds on GLUE benchmark with less than 1\% accuracy loss. Among the second line of research, various sparse attention techniques have been proposed, including local attention restricted to nearby tokens~\cite{Child2019GeneratingLS,Zaheer2020BigBT,Roy2020EfficientCS}, introducing global tokens~\cite{Guo2019StarTransformer,Zaheer2020BigBT}, content-based token grouping~\cite{kitaev2020reformer,Roy2020EfficientCS}, and attention back tracking~\cite{Lee2023SparseTT}. With the advance of the modeling capacity of LLM, the context input for LLM is also growing readily. To provide a more adaptive attention selection, Anagnostidis~\etal~\cite{Anagnostidis2023DynamicCP} proposes a sparse sigmoid-based selection scheme. Also, erasing tokens from the key-value cache makes this method more hardware-friendly in a decoder-based language model. While the experiments are mainly conducted on modest size GPT-2 model~\cite{Radford2019LanguageMA}, the empirical results that up to 80\% of the context can be successfully pruned with negligible degradation in perplexity indicates an appealing potential to explore along this direction.     

\subsubsection{Pruning for Transformer-Based Vision Model} 

\begin{table*}[t]
	\centering
	\small
	\renewcommand\arraystretch{1.0}
	\setlength{\tabcolsep}{6pt}
	\caption{\small{comparison of representative pruning methods on vision Transformer DeiT-Base. The Speed-up ratio is calculated using the latency or throughput reported in the original paper. Top-1 Acc. is measured on ImageNet 1K. ``/'' indicates the keep ratio of dynamic tokens, and ``-'' denotes the pruning ratio of parameters.}}
	\vspace{-0.5em}
	\begin{tabular}{l|c|c|c|c|c}
		\Xhline{1pt}
		Models & Pruning Type &  $\#$ Params & FLOPs & Speed Up & Top-1 Acc. (\%) \\		
		\hline
		DeiT-B	& - & 86.6M	& 17.6G	& 1$\times$	& 81.8\% \\
		
		\hline
		DPS-DeiT-B~\cite{Tang2021PatchSF} & Unstructured (Token) & 87M & 9.4G & 1.4$\times$ & 81.5\% \\
		DynamicVIT-B/08~\cite{Rao2021DynamicViTEV}	& Unstructured (Token) & -	& 13.3G & 1.2$\times$ & 81.6\% \\
		S$^2$ViTE-B~\cite{Chen2021ChasingSI} & Un- and Structured &	56.8M &	11.7G &	1.33$\times$ &	82.2\% \\
		SAViT-50~\cite{Zheng2022SAViTSV} & Structured &	42.6M &	8.8G	&	1.55$\times$ &	82.5\% \\
		SAViT-70~\cite{Zheng2022SAViTSV} & Structured & 25.4M &	5.3G	&	2.05$\times$ &	81.7\% \\
		ViT-Slim~\cite{Chavan2022VisionTS} & Structured & 52.6M & 10.6G	&	-	& 82.4\% \\
		WD-prining~\cite{Yu2022WidthD} & Structured and Early-Ex & 55.3M	& 9.9G	&	1.18$\times$ &	80.76\% \\
		X-pruner~\cite{Yu2023XPrunerEP} & Structured & 87M	$^\dagger$	& 8.5G	& - &	81.02\% \\
		\Xhline{1pt}
	\end{tabular}
	\label{table:pr-cv}
	\begin{tablenotes}
		\item $^\dagger$ The corresponding data is from~\cite{Papa2023ASO}.
	\end{tablenotes}
\end{table*}

\begin{table*}[t]
	\centering
	\small
	\renewcommand\arraystretch{1.0}
	\setlength{\tabcolsep}{6pt}
	\caption{\small{comparison of representative pruning methods on large language Transformer.}}
	\vspace{-0.5em}
	\begin{tabular}{l|c|c|c|c|c}
		\Xhline{1pt}
		\multirow{2}{*}{Method} & \multirow{2}{*}{Pruning Type} & \multirow{2}{*}{Source Model} &  Compression & Speed & Comparison with \\	
		&  &   &   Rate & Up & Source Model\\	
		\hline
		\multirow{2}{*}{SparseGPT~\cite{Frantar2023SparseGPTML}$^\dagger$}	& \multirow{2}{*}{Un- and semi-sturctured} & \multirow{2}{*}{OPT-175B~\cite{Zhang2022OPTOP} }&  \multirow{2}{*}{50-60\% }& \multirow{2}{*}{1.54-1.79$\times$} & 8.38/8.34  \\
		&  &   &   && (WikiText2 ppl)\\	 \hline
		\multirow{2}{*}{Sheared LLama~\cite{Xia2023ShearedLA}}	& \multirow{2}{*}{Structured} & \multirow{2}{*}{LLaMA2-7B~\cite{touvron2023llama}} & \multirow{2}{*}{61.4\% }& \multirow{2}{*}{-} & 56.7\%/64.6\%  \\
		&  &   &  && (11 datasets)\\	 \hline
		\multirow{2}{*}{LLM Pruner~\cite{Xia2023ShearedLA}}	& \multirow{2}{*}{Structured} &\multirow{2}{*}{ LLaMA2-7B~\cite{touvron2023llama}} & \multirow{2}{*}{20\%} & \multirow{2}{*}{1.18$\times$ }& 60.07\%/68.59\%  \\
		&  &   &  && (7 datasets)\\ \hline
		\multirow{2}{*}{Dyna. Context Pruning\cite{Anagnostidis2023DynamicCP}} & \multirow{2}{*}{Context} & \multirow{2}{*}{GPT-2~\cite{Radford2019LanguageMA}} & \multirow{2}{*}{80.35\%$^\star$} & \multirow{2}{*}{1.2$\times$} & +0.085 ppl\\
		&  &   &  && (Wiki and bookcorpus)\\
		\Xhline{1pt}
	\end{tabular}
	\begin{tablenotes}
		\item $^\dagger$ Data is from 2:4 partial sparsity setting with GPU speedup~\cite{Frantar2023SparseGPTML}.
		\item $^\star$ Compression rate refers to context tokens instead of parameters.
	\end{tablenotes}
	\label{table:pr-llm}
\end{table*}

\textbf{Token and feature pruning.} The early exploration of pruning for vision Transformer pruning focuses on reducing information redundancy in feature space. In a preliminary study, Zhu~\etal~\cite{Zhu2021VisionTP} propose learning the most informative feature dimensions and neglecting the remaining dimensions in the inference time. This is achieved by learning a diagonal real-value feature mask during pre-training with a sparsity regularizer and obtaining the hard feature mask using a defined threshold. From a different perspective, given the fact that token features in late ViT layers are quite similar in feature embedding, patch slimming~\cite{Tang2021PatchSF} method is proposed to prune the patches from the output layer to the input layer. In particular, for each layer, it evaluates the impact of a patch on the final output feature and removes less important ones. Besides determining the patch importance by training dataset statistics, token sparsification can also be achieved by inserting a light-weight prediction network to indicate the importance of the input patches~\cite{Tang2021PatchSF,Rao2021DynamicViTEV,Wang2021NotAI,Liang2022NotAP,Chen2021ChasingSI}. Note that since the self-attention can accept the token sequence regardless of its length, unstructured pruning of input tokens can be easily implemented in a hardware-friend manner for these token pruning methods. Furthermore, similar techniques are adopted in other Transformer-based vision models such as DETR~\cite{Carion2020EndtoEndOD} to promote focus on object tokens~\cite{Roh2021SparseDE,Zheng2023LessIM}. 

\textbf{Structured pruning.} Apart from the redundancy in feature dimension and tokens, the heavy burden of matrix computation is another important axis to improve model efficiency. In an early work,~\cite{Chen2021ChasingSI} proposes to selectively optimize for sparsity by obtaining unstructured subnetworks and structured pruning for attention heads, besides importance-based token selection.~\cite{Yu2022WidthD} develops a structural pruning method to reduce the network's width by pruning linear projection matrices as well as attention heads, together with depth pruning with inserted early-exit classifiers at each layer. Later on, more works emerged to explore the collaborative effect of parameter pruning across different Transformer modules. In~\cite{Yang2021GlobalVT}, a Hessian-based criteria is proposed to evaluate the parameter importance globally over learnable parameters. Together with the parameter redistribution strategy that promotes shorter latency, this method can also introduce novel efficient ViT structures named NViT. With an emphasis on speeding up continuous structure search, Chavan~\etal~\cite{Chavan2022VisionTS} perform a single-shot architecture search for all the ViT components with significantly fewer GPU hours needed. Also, by incorporating the interactions between the components, Zheng~\etal~\cite{Zheng2022SAViTSV} manage to introduce adaptively the pruning ratio for each component, adding another degree of freedom in pruning. More recently, Yu~\etal~\cite{Yu2023XPrunerEP} propose to use class labels to optimize the fully differentiable pruning operation at each layer.

\subsubsection{Discussion} Pruning serves as a fundamental way to reduce the computation burden of the pre-trained Transformer. Traditionally, this is achieved in an orthogonal manner by pruning input features or network parameters. In particular, we notice that different pruning methods can biasedly benefit particular application purposes. For large language models, structured network pruning fits more in the setting when a modest length (e.g., several thousand tokens) is used as input to speed up inference, as the defined hardware-friendly model shape has been proven to realize a good trade-off between acceleration and perplexity decrease~\cite{Xia2023ShearedLA}. In comparison, context pruning or sparse attention methods deserve more attention when context length gets longer, and the ceiling of pruning ratio for context for LLMs (e.g., 7B or more) has not been determined in a detailed manner, which is an interesting direction to explore. For vision Transformers, the development of pruning methods has systematically tackled each angle of model design, making the current pruning process more flexible and closer to a continuous structural search task, which can be an excellent reference to the development of language domains. The training cost is another critical axis to consider in the pruning of large models, especially when incremental LLM pre-training is considered a component of the pruning process. Therefore, developing training-efficient pruning methods is attracting more attention. This requires the parameter (or block) sensitivity to be estimated more accurately using limited data or the information stored in hidden features during training/inference to be explored more in-depth.

\subsection{Efficient Architecture}
\label{sec:effi}

Transformer~\cite{vaswani2017attention} has become the de facto backbone architecture to develop various large vision and language models (LVMs and LLMs), making it possible to scale to billions of parameters. In this section, we review the efficient architecture design of large Transformer-based models in both vision and language domains, \ie, efficient design on mainstream architecture, attention mechanism and feed forward network.
Notably, here we also discuss the recently proposed Mamba~\cite{gu2023mamba}, a general sequence model backbone that integrates selective state space models (SSMs) into a simplified end-to-end neural network architecture. In this\\
	 section, we consider it as an efficient design for subquadratic-time architectures.

\subsubsection{Representative Transformer-Based Models}
Table~\ref{table:efficient_arch} presents the model cards of several representative LLMs and LVMs with public details.
Vanilla Transformer~\cite{vaswani2017attention} is based on the encoder-decoder architecture~\cite{raffel2020exploring}. In this architecture, the encoder stacks multi-head self-attention layers to encode the input sequence into latent representations, while the decoder performs cross-attention on these representations and generates target sequences in an autoregressive manner. BERT~\cite{devlin2018bert} is designed to pre-train deep bidirectional representations from unlabeled text by considering both left and right context in all layers.
GPT-series models~\cite{brown2020language,gpt4} have effectively demonstrated the power of in-context learning using decoder-only architectures~\cite{scao2022bloom,rae2021scaling,penedo2023refinedweb}. These decoder architectures incorporate unidirectional attention masks to ensure that each input token can only attend to previous tokens, including itself. The input and output tokens are processed in a similar manner within the decoder.
In the following, we will provide separate reviews from the perspectives of CV and NLP on the current structural innovations in  Transformer-based architectures.

\subsubsection{Language Domain} 
The self-attention mechanism in conventional Transformer architectures often faces quadratic computational complexity. This poses a challenge for training and making inferences with long input sequences. To enhance efficiency, current innovations in the mainstream can be classified into three categories: optimizations targeting the attention mechanism, the direct replacement of attention with more efficient architectures, and enhancements focused on the FFN component.

\noindent\textbf{Attention innovation.} 
Various efficient Transformer variants have been proposed to reduce the computational complexity of the attention mechanism by incorporating structural priors on attention, such as \emph{sparsity}~\cite{peng2021random,zaheer2020big,child2019generating,roy2021efficient,rae2019compressive,xiao2023efficient}. For example, Reformer~\cite{kitaev2020reformer} approximates full attention computation using locality-sensitive hashing, reducing the complexity from $\mathcal{O}(N^2)$ to $\mathcal{O}(N\log N)$. Other techniques include pooling-based compression~\cite{rae2019compressive}, clustering methods~\cite{roy2021efficient,khandelwal2019generalization} that apply k-means clustering to learn dynamic sparse attention regions, and Longformer~\cite{beltagy2020longformer}, which combines local windowed attention with task-motivated global attention.
Chelba~\etal~\cite{chelba2020faster} proposes truncating the target-side window based on an $N$-gram assumption, restricting the self-attention mechanism to use only the previous $N-1$ tokens. Additionally, locally banded sparse attention methods, such as Factorized Attention~\cite{child2019generating}, have been adopted in models like GPT-3~\cite{brown2020language}.

Other methods have been proposed to reduce the computational complexity of the attention mechanism using \emph{linear attention}.
Performer~\cite{choromanski2020rethinking} introduces fast attention via positive orthogonal random features, which is a scalable kernel method for approximating softmax attention kernels. This approach can efficiently model kernelizable attention mechanisms beyond softmax and comes with strong theoretical guarantees, including nearly unbiased estimation of the attention matrix, uniform convergence, and low estimation variance.
Linear Transformer~\cite{katharopoulos2020Transformers} reformulates self-attention as a linear dot-product of kernel feature maps. It uses kernels $\frac{\phi(q_i)\phi(k_j)}{\sum_{n=1}^{|x|}\phi(q_i)\phi(k_n)}$ to replace the softmax function and takes advantage of the associativity property of matrix products to reduce the computational complexity from $\mathcal{O}(N^2)$ to $\mathcal{O}(N)$.

Moreover, there have been studies~\cite{chowdhery2022palm,li2023starcoder,shazeer2019fast} exploring the concept of \emph{multi-query attention}, where different heads share the same linear transformation matrices for the keys and values, as proposed by Shazeer \etal\cite{shazeer2019fast}. This approach offers significant computational savings with only a minimal impact on model quality. To further strike a balance between multi-query attention and multi-head attention, GQA~\etal~\cite{ainslie2023gqa} introduces the concept of grouped attention heads, where heads within the same group share identical transformation matrices.

\begin{table}[t]
	\centering
	\footnotesize
	\renewcommand\arraystretch{1.0}
	\setlength{\tabcolsep}{3.8pt}
	\caption{\small{Model cards of several representative Transformer-based LLMs and LVMs with public configuration details. Here $d_{model}$ is already expanded by expansion ratio.}}
	\vspace{-0.5em}
	\begin{tabular}{l|c|c|c|c|c}
		\Xhline{1pt}
		Models & Arch. & $\#$ Params & Layer & Head & $d_{model}$ \\
		\hline
		\multicolumn{6}{l}{Natural Language Processing} \\
		\hline
		BERT~\cite{devlin2018bert} & Encoder & 0.1B-0.3B & 12-24 & 12-16 & 3072-4096 \\
		PaLM~\cite{chowdhery2022palm} & Decoder & 9B-540B & 32-118 & 16-48 & 4096-18432 \\
		Gopher~\cite{rae2021scaling} & Decoder & 4M-280B & 8-80 & 16-128 & 512-16384 \\
		BLOOM~\cite{scao2022bloom} & Decoder & 0.6B-176B & 24-70 & 16-112 & 1024-14336 \\
		GPT3~\cite{brown2020language} & Decoder & 175B & 96 & 96 & 12288 \\
		GLM-130B~\cite{zeng2022glm} & Decoder & 130B & 70 & 96 & 12288 \\
		Chinchilla~\cite{hoffmann2022training} & Decoder & 70B & 80 & 64 & 8192 \\
		LLaMA~\cite{touvron2023llama} & Decoder & 7B-65B & 32-80 & 32-64 & 4096-8192 \\
		Falcon~\cite{penedo2023refinedweb} & Decoder & 40B & 60 & 64 & 8192 \\
		T5~\cite{raffel2020exploring} & En.-De. & 11B & 24 & 128 & 1024 \\
		\hline
		\multicolumn{6}{l}{Computer Vision} \\
		\hline
		ViT~\cite{dosovitskiy2020image} & Encoder & 86M-0.6B & 12-32 & 12-16 & 3072-5120 \\
		DeiT~\cite{touvron2021training} & Encoder & 5M-86M & 12 & 3-12 & 768-3072 \\
		TNT~\cite{han2021Transformer} & Encoder &6M-66M & 12 & 3-10 & 768-2560 \\
		PVT~\cite{wang2021pyramid} & Pyramid & 13M-61M & 12-45 & 8 & 2048 \\
		CMT~\cite{guo2022cmt} & Pyramid &10M-75M & 24-40 & 8 & 1324-2432 \\
		Swin~\cite{liu2021swin} & Pyramid & 29M-0.2B & 16-28 & 24-48 & 3072-6144 \\
		\Xhline{1pt}
	\end{tabular}
	\label{table:efficient_arch}
\end{table}

FlashAttention~\cite{dao2022flashattention} offers an innovative approach to optimizing the speed and memory utilization of attention modules on GPUs. It focuses on an Input/Output (IO)-aware perspective, making better use of the fast memory SRAM by dividing the input into blocks. This optimization technique has already been integrated into various platforms~\cite{rasley2020deepspeed,shoeybi2019megatron,paszke2019pytorch,li2023colossal,wolf2019huggingface} developed for training LLMs.

\noindent\textbf{Non-Transformer architecture.}
Recent research has focused on developing new architectures for language modeling, including parameterized state space models~\cite{gu2021efficiently,mehta2022long,dao2022hungry,gupta2022diagonal}, long convolutions~\cite{poli2023hyena}, and the incorporation of recurrent models~\cite{dai2019Transformer,martins2021infty,sun2023retentive,ma2022mega}, which come with strong inductive biases.
In particular, Gu~\etal~\cite{gu2021efficiently} introduces the S4 model, based on a novel parameterization for the SSM~\cite{chen1984linear,gu2020hippo}, which involves conditioning matrix A with a low-rank correction, making it stably diagonalizable. This reduces the SSM to the well-studied computation of a Cauchy kernel~\cite{pan2017fast}. The GSS model~\cite{mehta2022long} further extends the use of gating units to the state space model family, and it is observed that employing gating units leads to dimensionality reduction when performing FFT operations, which addresses a major bottleneck in training speed.
Hyena~\cite{poli2023hyena} trains a recurrence of gating units and implicitly parametrized long convolutions, which serves as an attention-free drop-in replacement for the traditional Transformer architecture.
Ma~\etal~\cite{ma2022mega} introduces MEGA, a single-head gated attention mechanism enhanced with exponential moving average. This mechanism aims to integrate the inductive bias of position-aware local dependencies into the position-agnostic attention mechanism.

Peng~\etal~\cite{peng2023rwkv} proposes RWKV based on the fact that Recurrent Neural Networks (RNNs) show linear scaling in memory and computational requirements but encounter limitations in parallelization and scalability. They leverage a linear attention mechanism, specifically replacing the quadratic QK attention with a scalar formulation that has linear cost. Additionally, they redefine recurrence and sequential inductive biases to parallelize computations during training while maintaining constant computational and memory complexity during inference.
\begin{figure}[t]
	\centering
	\begin{subfigure}[b]{0.22\textwidth}
		\centering
		\includegraphics[width=0.46\textwidth]{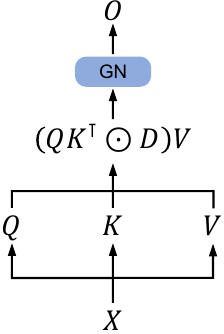}
		\caption{Parallel representation.}
		\label{fig:retnet:parallel}
	\end{subfigure}
	\hspace{0.1pt}
	\begin{subfigure}[b]{0.24\textwidth}
		\centering
		\includegraphics[width=\textwidth]{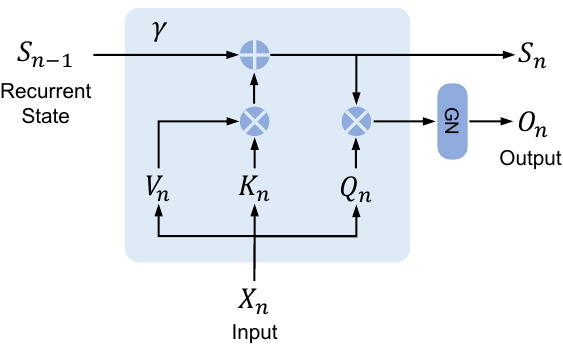}
		\caption{Recurrent representation.}
		\label{fig:retnet:recurrent}
	\end{subfigure}
	\caption{\small{Dual form of RetNet (image from~\cite{sun2023retentive}).}}
\end{figure}
Furthermore, Retentive Network (RetNet)~\cite{sun2023retentive} theoretically derives the connection between recurrence and attention and proposes the retention mechanism for sequence modeling. This mechanism supports three computation paradigms, namely parallel, recurrent, and chunkwise recurrent, achieving training parallelism, low-cost inference, and good performance simultaneously. The parallel representation (also depicted in Figure~\ref{fig:retnet:parallel}) of retention is:
\begin{equation}
	\begin{aligned}
		\label{eq:ret:parallel}
		Q = (X W_Q) \odot \Theta ,&\quad K = (X W_K) \odot \overline{\Theta} ,\quad V = X W_V \\
		\Theta_n = e^{in\theta},& \quad
		D_{nm} =
		\left\{
		\begin{aligned}
			& \gamma^{n-m}, &n\ge m \\
			& 0, &n < m \\
		\end{aligned}
		\right.
		\\
		\mathrm{Rete}&\mathrm{ntion}(X) = (Q K^\intercal \odot D)V
	\end{aligned}
\end{equation}
where $\overline{\Theta}$ is the complex conjugate of $\Theta$, and $D \in \mathbb{R}^{|x|\times |x|}$ combines causal masking and exponential decay along relative distance as one matrix.
This parallel representation enables RetNet to train the models with GPUs efficiently.
The recurrent version of retention (Figure~\ref{fig:retnet:recurrent}) is favorable for inference and can be written as (for $n$-th timestep):
\begin{equation}
	\begin{aligned}
		\label{eq:ret:recurrent}
		&S_n = \gamma S_{n-1} + K_n^{\intercal} V_n \\
		&\mathrm{Rete}\mathrm{ntion} (X_n) = Q_n S_n, \quad n = 1, \cdots, |x| \\
	\end{aligned}
\end{equation}
where $Q,K,V,\gamma$ are the same as in Eq.~\ref{eq:ret:parallel}.

More recently, Gu~\etal~\cite{gu2023mamba} have highlighted the limitations of previous structured SSMs in effectively modeling discrete and information-dense data, such as text. In response, they propose Mamba, consisting a novel class of selective mechanisms and hardware-aware designs that enable linear scalability to billions of parameters, while maintaining computational efficiency and achieving strong performance. Specifically, in comparison to the H3~\cite{dao2022hungry} block, the Mamba model replaces the first multiplicative gate with an activation function. Additionally, Mamba incorporates an SSM into the main branch, distinguishing it from the MLP block in transformer. Building upon the foundation of Mamba, researchers are further unlocking the potential of SSMs for scaling by exploring new techniques, such as combining SSMs with Mixture of Experts (MoE)~\cite{pioro2024moe} or introducing methods to enhance the flow of hidden information~\cite{he2024densemamba} between layers in SSMs.

\noindent\textbf{FFN innovation.}
Large and sparse FFNs such as Mixture-of-Experts (MoE)~\cite{du2022glam,lepikhin2020gshard,fedus2022switch,roller2021hash,chi2022representation,lewis2021base} have been effective in scaling up Transformer-based models for pre-training LLMs. They replace a single FFN module with multiple equally-sized modules (experts) and activate only a few experts based on the input. This selective activation of FFN improves generalization performance while maintaining fixed training and inference costs.
Liu~\etal~\cite{liu2023towards} further finds a simpler selection method known as Avg-K, which selects blocks based on their mean aggregated hidden states. This method achieves lower perplexity in language model pre-training compared to existing MoE architectures such as Switch Transformer~\cite{fedus2022switch} and HashLayer~\cite{roller2021hash}.

Romero~\etal~\cite{romero2021ckconv} presents Continuous Kernel Convolution (CKConv) for handling arbitrarily long sequences in a parallel manner within a single operation. CKConv formulates convolutional kernels as vector-valued continuous functions parameterized by a small MLP instead of a sequence of independent weights. This MLP takes a time-step as input and generates the value of the convolutional kernel at the corresponding position, allowing the generation of kernels at different resolutions and arbitrarily sizes.

PaLM~\cite{chowdhery2022palm} and LLaMA~\cite{touvron2023llama} both utilize the SwiGLU Activation (Swish($xW$) $\dot xV$) for original FFN intermediate activations. This choice is based on the observation that SwiGLU activations have been shown to substantially enhance quality in compute-equivalent experiments~\cite{shazeer2020glu} when compared to standard activation functions such as ReLU, GeLU, or Swish~\cite{shazeer2020glu}. It's important to note that using SwiGLU does require three matrix multiplications in the MLP instead of two.

\begin{table}[t]
	\footnotesize
	\setlength{\tabcolsep}{2pt}
	\caption{\small{Model comparison from training parallelization (TP), inference cost (Time), and memory complexity (Memory). Here $N$ and $d$ denote the sequence length and feature dimension, respectively.}}
	\label{table:efficient_llm_compare}
	\begin{tabular}{lccc}
		\toprule
		Architectures      & TP & Time & Memory \\
		\midrule
		RNN~\cite{rumelhart1986learning}        & \ding{56} & $O(Nd)$ & $O(d)$  \\
		Transformer~\cite{vaswani2017attention} & \ding{52} & $O(N^2d)$ & $O(N^2 + Nd)$  \\
		Linear Attention~\cite{katharopoulos2020Transformers} & \ding{52} & $O(Nd^2)$ & $O(Nd + d^2)$  \\
		H3~\cite{dao2022hungry}              & \ding{52} & $O(Nd(\log N+d))$ & $O(Nd)$  \\
		S4~\cite{gu2021efficiently}          & \ding{52} & $O(Nd^2)$ & $O(Nd)$  \\
		Hyena~\cite{poli2023hyena}           & \ding{52} & $O(Nd(\log N+d))$ & $O(N\log N\cdot d)$  \\
		Reformer~\cite{kitaev2020reformer}   & \ding{52} & $O(N\log N\cdot d)$ & $O(N\log N + Nd)$ \\
		Performer~\cite{choromanski2020rethinking} & \ding{52} & $O(Nd^2\log d)$ & $O((Nd + d^2)\cdot\log d)$ \\
		RWKV~\cite{peng2023rwkv}             & \ding{52} & $O(Nd)$ & $O(d)$  \\
		RetNet~\cite{sun2023retentive}       & \ding{52} & $O(Nd^2)$ & $O(d^2)$ \\
			Mamba~\cite{gu2023mamba}       & \ding{52} & $O(Nd)$ & $O(d)$ \\
		\bottomrule
	\end{tabular}

\end{table}

\subsubsection{Vision Domain}
Dosovitskiy~\etal~\cite{dosovitskiy2020image} first introduces the Vision Transformer (ViT) primarily designed for image classification tasks. In the ViT model, the input image is initially divided into a sequence of fixed-length tokens, which are then processed through multiple Transformer layers to capture global relationships among the tokens.
Similar to the challenges faced in the language domain, the self-attention mechanism with quadratic computational complexity is one of the major hurdles for training and inference in Transformer-based architectures applied to vision tasks. Many dense prediction tasks, such as object detection and image segmentation, often require higher-resolution inputs, necessitating the manual design of efficient architectures to handle these challenges.
To address these issues, novel architectures like PVT~\cite{wang2021pyramid} and Swin~\cite{wang2021pyramid} have been proposed. These architectures incorporate hierarchical (pyramid) structures with multiple stages, effectively overcoming the challenges of adapting the original isotropic ViT~\cite{dosovitskiy2020image} to various dense prediction tasks in computer vision.

In the following, we will delve into papers that focus on improving the efficiency of ViTs, including efforts to enhance local information processing~\cite{guo2022cmt,heo2021rethinking,yang2021focal,tang2022quadtree,pan2021scalable,chen2021crossvit,xu2021co}, simplify the attention mechanism~\cite{parmar2018image,hatamizadeh2023fastervit,liu2023efficientvit,you2022castling,xiong2021nystromformer}, and explore alternative modules that can replace or work alongside attention mechanisms~\cite{guo2022beyond,yu2022metaformer,wang2023riformer,lian2021mlp,chen2021cyclemlp}, \etc.

\noindent\textbf{Enhancing locality.}
T2T-ViT~\cite{yuan2021tokens} leverages token transformations to reduce token length by iteratively aggregating neighboring tokens to one token. 
LeViT~\cite{graham2021levit} and MobileViT~\cite{mehta2021mobilevit} employ hybrid architectures with stacked convolution layers, efficiently reducing the number of features through the first layer.
TNT~\cite{han2021Transformer} further divides original ViT's 16$\times$16 patches into smaller 4$\times$4 patches for capturing local information.
CrossViT~\cite{chen2021crossvit} processes small-patch and large-patch tokens separately, fusing them through multiple attention mechanisms.
Twins~\cite{chu2021twins} alternates between local and global attention layers for improved performance.
RegionViT~\cite{chen2021regionvit} introduces regional tokens and local tokens, enhancing local context with global information.
KVT~\cite{wang2022kvt} introduces the k-NN attention to utilize locality of images patches and ignore noisy tokens by only computing attentions with top-k similar tokens.
CMT~\cite{guo2022cmt} applies depth-wise convolutions to augment local patterns in the attention map and intermediate activation of the FFN.
Pan~\etal~\cite{pan2022fast} proposes HiLo attention to disentangle the high/low frequency patterns in an attention layer by separating the heads into two groups, each equipped with specialized operations that focus on local windows.

\noindent\textbf{Faster attention.}
Parmar~\etal~\cite{parmar2018image} introduces restrictions on the attention mechanism to focus on local neighborhoods. Swin~\cite{liu2021swin} and Cswin~\cite{dong2022cswin} incorporate local attention within a window and introduce a shifted window partitioning method to enable cross-window connections. Shuffle Transformer~\cite{huang2021shuffle} and Msg-Transformer~\cite{fang2022msg} employ spatial shuffle operations as alternatives to shifted window partitioning, facilitating cross-window connections. FasterViT~\cite{hatamizadeh2023fastervit} introduces hierarchical attention, breaking down global self-attention into multi-level attention components. This approach combines local window attention and hierarchical attention to achieve global information propagation while reducing computational costs.
FLatten Transformer~\cite{han2023flatten} integrates depth-wise convolution in conjunction with linear attention mechanisms to address the challenge of maintaining diversity in output features across different positions. 

\noindent\textbf{Attention-free architecture.}
AFT~\cite{zhai2021attention} pioneers an innovative approach by combining key and value elements with learned position biases, followed by element-wise multiplication with the query. This operation's distinctive feature is its linear memory complexity concerning both context size and feature dimension, making it compatible with large input and model sizes.
GFnet~\cite{rao2021global} presents an alternative by replacing traditional attention mechanisms with Fast Fourier Transform (FFT), frequency gating, and inverse FFT for rapid token mixing.
Additionally, some architectures have been entirely based on pure multi-layer perceptrons (MLPs)~\cite{touvron2022resmlp,tolstikhin2021mlp,guo2022hire,tang2022image,lian2021mlp,chen2021cyclemlp}, without using convolutions or self-attention mechanisms.
Yu~\etal~\cite{yu2022metaformer} proposes MetaFormer as a general architecture abstracted from ViTs without specifying the token mixer. By employing basic token mixing, primarily through non-parametric pooling, MetaFormer achieves satisfactory performance.

Inspired by the efficient hardware-aware designs of Mamba in NLP, Zhu~\etal~\cite{zhu2024vision} propose a novel generic vision backbone called Vim with bidirectional Mamba blocks. 
Different from the sequential nature of language, representing visual data in SSMs presents challenges primarily due to their sensitivity to spatial position. To address this, Vim introduces position embeddings to mark image sequences and compresses visual representations using bidirectional SSM. Following the architectures of ViT~\cite{Dosovitskiy2020AnII} and DeiT~\cite{Touvron2020TrainingDI}, Vim initially employs a 16×16 kernel size projection layer to obtain a 1-D sequence of non-overlapping patch embeddings.
Simultaneously, Liu~\etal~\cite{liu2024vmamba} propose the visual state space model (VMamba), which introduces the Cross-Scan Module (CSM) to traverse the spatial domain and convert any non-causal visual image into ordered patch sequences, thereby addressing the inherent direction-sensitive issue. Following the architecture of Swin~\cite{Liu2021SwinTH}, VMamba begins the process by partitioning the input image into patches using a stem module, without further flattening the patches into a 1-D sequence.

\noindent\textbf{NAS.}
Indeed, the quest for optimizing Transformer architectures extends to the realm of neural architecture search (NAS). Various models, including Scaling-ViT~\cite{zhai2022scaling}, ViTAS~\cite{su2022vitas},
AutoFormer~\cite{chen2021autoformer} and GLiT~\cite{chen2021glit}, have emerged as products of NAS-driven design, demonstrating the potential for more efficient and effective Transformer architectures. 

\subsubsection{Discussion}

Transformer models consistently achieve state-of-the-art results across various tasks, but the cost of training these models, especially on long sequences, can be prohibitive. Table~\ref{table:efficient_llm_compare} concludes several papers that propose improvements to the attention mechanism from three different dimensions. While using linear attention can help reduce the computational complexity of the attention mechanism, it struggles to effectively encode position information, leading to a potential decrease in model performance.
Models like RWKV~\cite{peng2023rwkv}, RetNet~\cite{sun2023retentive} and Mamba~\cite{gu2023mamba}, which generate outputs recursively similar to RNNs, present a promising direction for further exploration. These models only need to refer to the previous state during decoding, significantly improving decoding efficiency by avoiding the need to revisit all previous states, a characteristic of conventional Transformers. Additionally, they can encode entire sentences in parallel, enabling highly parallel and efficient training.

In contrast to NLP tasks, high-level vision tasks place a strong emphasis on the network's ability to capture local details, contextual information, and multi-scale features for dense predictions~\cite{wang2021pyramid}. Several studies~\cite{chen2023vanillanet,yu2022metaformer} have also demonstrated that the self-attention mechanism is not always indispensable for feature extraction in vision tasks. In the exploration of pure vision tasks, the possibility of directly omitting the attention module is a worthwhile avenue for further research. However, it's worth noting that attention and cross-attention modules still play a crucial role in integrating visual features with other modalities~\cite{radford2021learning}. Therefore, the quest for faster attention algorithms remains a valuable research direction.

\section{Specialized Approaches}
\label{sec:other}
In addition to quantization, distillation, pruning, and novel network architectures, there are several other model compression and acceleration approaches including tensor decomposition, early exiting, and speculative sampling.

\noindent\textbf{Tensor decomposition.}
Tensor or matrix decomposition aims to decompose a large tensor or matrix into smaller ones in order to reduce the number of parameters and computational costs. This approach was first introduced into the compression of fully-connected layers and convolutional networks~\cite{denil2013predicting,jaderberg2014speeding}. In terms of large language models, Tensor decomposition is utilized for simplifying model weight or embedding layers. 
Edalati \emph{et.al}~\cite{edalati2022kronecker} is the first attempt to use Kronecker decomposition
for compressing generative language models. This work represent weight matrix with two smaller matrices using Kronecker decomposition and achieves lossless $1.5\times$ compression of GPT2. LoRD~\cite{kaushal2023lord} further uses low rank decomposition to compress code LLMs. And TSVD~\cite{chen2023ternary} compresses linear mappings in Transformers using SVD and constrain $U$ and $V$ matrices in ternary format. In addition to model weights, TensorGPT~\cite{xu2023tensorgpt} proposes to compress embedding layers and achieves more than $3\times$ compression factor with lossless performance.

\noindent\textbf{Early exiting.}
Early exiting can dynamically allocate different resources for each input sample and maintain the original performance. This technology has been successfully used in information retrieval system~\cite{cambazoglu2010early} and convolutional networks~\cite{teerapittayanon2016branchynet}. Many early exiting technologies have been proposed for encoder-only Transformers~\cite{stickland2019bert,schwartz2020right,liu2020fastbert,hou2020dynabert}. The key problem in early exiting is determining when to exit. The existing works mainly utilize intrinsic confidence measures~\cite{stickland2019bert}, routing in advance~\cite{liu2021faster}, or training a early-exit classifier~\cite{schuster2021consistent}. For Transformers with decoders, several works have been proposed. For example, Elbayad \emph{et.al}~\cite{elbayad2020depth} introduces the depth-adaptive Transformer for accelerating machine translation. 
CALM~\cite{schuster2022confident} proposes to dynamically allocate different computational resources for different input and timesteps, achieving speedups of up to $3\times$ on T5 encoder-decoder model. Additionally, SkipDecoder~\cite{del2023skipdecode} bypasses tokens in lower layers to middle layers in order to enable batch inference and reuse KV caching, achieving inference speedups of $2\times$ to $5\times$ on OPT models. 

\noindent\textbf{Speculative sampling.}
Speculative sampling, a special acceleration approach for Transformer decoding, computes several tokens in parallel~\cite{leviathan2023fast,chen2023accelerating}. In large language models, decoding $K$ tokens requires $K$ runs of the model -- a slow process. Taking the reference tokens generated from smaller models, speculative sampling runs these tokens in parallel to significantly accelerate the decoding process. Moreover, the rejection scheme~\cite{chen2023accelerating} can preserve the distribution of the original LLM so that speculative sampling is theoretically lossless. For example, Yang \emph{et.al}~\cite{yang2023inference} further take the input text as reference tokens without introducing extra models. And LLMCad~\cite{xu2023llmcad} introduces an on-device inference engine for LLMs with speculative sampling as a key technology.

\noindent\textbf{Discussion.}
This section has reviewed the methods that differ from those mentioned earlier, covering tensor decomposition, early exiting, and speculative sampling. These methods do not change the basic operations but instead accelerate model inference by modifying the network connections. Tensor decomposition approximates the original weights with low-rank ones, potentially leading to performance degradation. Early exiting makes predictions by adaptively allocating resources on a per sample basis. And speculative sampling considers the specialization of decoders and can accelerate decoding in a lossless manner. These special methods inspire future development of new algorithms for accelerating LLMs or LVMs.

\section{Conclusions and Future Directions}
\label{sec:con}

In this survey, we systematically investigate the  compression methods for Transformer models. Compared with compression method for conventional models, there are unique considerations when compressing  Transformer models.  Unlike other architectures such as CNN or RNN, the Transformer adopts a unique architecture design with alternative attention and FFN modules. This necessitates specifically tailored compression methods for optimal compression rates. Moreover, the efficiency of compression methods becomes particularly crucial for  large models. Certain model compression methods require substantial amounts of computational resources, making them potentially prohibitive for such models. This survey aims to encompass the majority of recent works pertaining to  Transformers and articulate a comprehensive roadmap for their compression.  Subsequently, we delve into the interconnections among various methods, address post-challenges, and outline directions for future research.

\noindent\textbf{Relationship between different compression methods.}
Different compression methods can be used together to obtain an extremely efficient architecture.  A conventional sequence is to firstly define a new  architecture with efficient operations. Any redundant components (\eg, attention head, layers) are then removed to  reduce the model size. For practical hardware implementation, quantizing weights or activations to lower bits is imperative. The choice of required bits depend on not only the tolerance of error, but also  hardware design. As an illustration, Int8 computation is efficiently optimized on Nvidia A100 GPU, but it lacks support on the older Tesla P100 GPU. Distillation commonly serves as a training strategy that applies during the fine-tuning phases of both pruning and quantization. Combining different compression strategies is a promising direction to explore for achieving extremely high compression rates. Although detailed exploration has been conducted into conventional models like CNN~\cite{han2015deep,wang2020apq},  Transformer models have more complicated architectures and higher computational costs. Finding a suitable combination strategy via joint search is a challenging task.

\noindent\textbf{Training-efficient compression strategy.} 
In contrast to compressing conventional models, there is a heightened emphasis on the computational cost of compression methods. 

Different from compression for conventional models, the  cost of compressing large Transformers  has been put on higher priority. Large Transformers are currently trained on vast datasets using significant computational resources. For instance, Llama2 is trained on 2 trillion tokens with thousands of GPUs over several months \cite{touvron2023llama}. It is impractical to fine-tune a model with comparable computational resources during pre-training, especially when the original data is often inaccessible. Therefore, the feasibility of efficient compression methods after training becomes more viable.  

Initially developed for conventional small models, a series of works have extensively researched post-training quantization~\cite{liu2021post, yuan2022ptq4vit,lin2021fq,ding2022towards}, and these methods have seamlessly transitioned to  Transformers. With only several GPU hours, some recent works GPTQ~\cite{frantar2022gptq},and SmoothQuant~\cite{xiao2023smoothquant} quantize an FP16 model to Int8 without significant performance loss.  However,  for lower bits (\eg, 4 bit), the quantized model still surfers significant performance degradation~\cite{frantar2022gptq,xiao2023smoothquant}. It is worth noting that extremely-low-bit models, such as binary Transformers, have been extensively explored in conventional small models,  yet they remain relatively unexplored in the context of large models.

For pruning,  the challenge of post-training is intricately linked to the pruning granularity. Although nonstructural sparsity can achieve a high compression rate with minimal fine-tuning requirements~\cite{Frantar2023SparseGPTML}, a similar strategy is hard to transfer to the structural pruning. Directly removing the entire attention heads or layers will result in a substantial alteration of the model's architecture and a  significant reduction in accuracy.  How to recognize effective weights and how to effectively recover the performance are both important directions for exploration.
The efficient strategy to  identify effective weights and recovery representation ability are key research directions in addressing these challenges.

\noindent\textbf{Efficient architectures beyond Transformer.}
In real-world applications, the input context for a Transformer architecture can be extremely long, encompassing sequence texts (e.g., a book with hundreds of thousands of words) in NLP or high-resolution images in CV. The vanilla attention mechanism exhibits quadratic complexity regarding the length of the input sequence, posing a significant computational challenge for long-sequence inputs.  Numerous studies have addressed this issue by mitigating the computational cost of attention, employing techniques such as sparse attention, local attention, \etc (See Section \ref{sec:effi}). However, these attention compression strategies often compromise the representation ability, leading to diminished performance.

Emerging architectures such as RWKV~\cite{peng2023rwkv} and RetNet~\cite{sun2023retentive}  adopt a recursive output generation akin to RNNs, effectively reducing computational complexity to $\O(N)$. This development holds promise for further exploration in the quest for more efficient models.  For computer vision tasks, even a pure MLP architecture without an attention module can achieve SOTA performance~\cite{tolstikhin2021mlp,chen2021cyclemlp,tang2022image,touvron2022resmlp,guo2022hire}. Beyond the widely used Transformer architecture, it is promising to explore new efficient architectures by carefully investigating their efficiency, generalization and scalability. 


\renewcommand\refname{References}
{\small
	\bibliographystyle{unsrt2authabbrvpp}
	\bibliography{ref}
}

\end{CJK}
\end{document}